\documentclass[11pt]{article}
\usepackage[preprint]{acl}
\usepackage[hidecomments]{nert}
\usepackage{times}
\usepackage{latexsym}
\usepackage{graphicx} % Required for inserting images
\usepackage{linguex}

\usepackage{coptic}
\usepackage{tcolorbox}
\usepackage{xcolor}

\usepackage[export]{adjustbox}

\newcommand{\lexicon}[0]{\texttt{LEX}}
\newcommand{\dep}[0]{\texttt{DEP}}
\newcommand{\con}[0]{\texttt{CON}}
\newcommand{\conll}[0]{\texttt{CoNLLU}}
\newcommand{\syntax}[0]{\texttt{SYN}}
\newcommand{\all}[0]{\texttt{\lexicon+\syntax}}

\newcommand{\dev}[0]{\texttt{dev}}
\newcommand{\test}[0]{\texttt{test}}

\newcommand{\chrf}[0]{\texttt{chrF++}}

\definecolor{papyrusdark}{RGB}{219, 200, 172}
\newtcolorbox{mybox}{
  colback=gray!5,          % Very light background
  colframe=papyrusdark,  % Dark blue accent
  arc=6pt,                 % Sharp corners
  outer arc=6pt,
  boxrule=1pt,             % No outer frame
}

\title{Syntax as a Rosetta Stone: Universal Dependencies for In-Context Coptic Translation}

\author{
 \textbf{Abhishek Purushothama\footnotemark[1]}  \quad
 \textbf{Emma Thronson}\thanks{Equal contribution}  \quad
 \textbf{Alexia Guo}  \quad
 \textbf{Amir Zeldes}
\\
 Corpling Lab
 \\
 Georgetown University
 \\
 \{\emldisplay{ap2089@georgetown.edu}{ap2089}, \emldisplay{et726@georgetown.edu}{et726}, \emldisplay{qg65@georgetown.edu}{qg65}, \emldisplay{amir.zeldes@georgetown.edu}{amir.zeldes}\}\texttt{@georgetown.edu}
}
\date{April 2026}

\begin{document}

\maketitle
\begin{abstract}
Low-resource machine translation requires methods that differ from those used for high-resource languages.
This paper proposes a novel in-context learning approach to support low-resource machine translation of the Coptic language to English, with syntactic augmentation from Universal Dependencies parses of input sentences. Building on existing work using bilingual dictionaries to support \text{inference} for vocabulary items, we add \text{several representations} of syntactic analyses to our \text{inputs}, specifically exploring the inclusion of raw parser outputs, verbalizations of parses in plain English, and targeted instructions of difficult constructions identified in sub-trees and how they can be translated. Our results show that while \text{syntactic information} alone is not as useful as dictionary-based glosses, combining retrieved dictionary items with \text{syntactic information} achieves significant gains across model sizes, achieving new state-of-the-art translation results for Coptic.
\end{abstract}

\section{Introduction}

Recent advances in LLMs have raised the quality of baseline results of machine translation (MT) in high-resource languages (HRLs) to the point where they can be used in user facing and downstream application contexts \cite{zhu-etal-2024-multilingual}. At the same time, low-resource languages (LRLs) have seen limited benefits from baseline prompting approaches \cite{frontull2025compensatingdatareasoninglowresource, PavaEtAl2025MindGap}, since models have little or no language modeling capabilities for low-resource languages.

\begin{figure}[t!]
\begin{mybox}
\includegraphics[width=\columnwidth, frame]{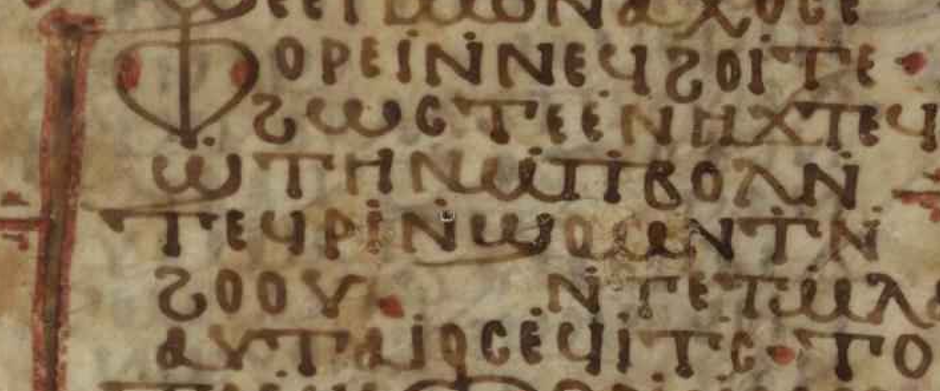}\label{img:excerpt-photo}
     An excerpt from Apophthegmata Patrum 25, MS MONB.EG 67 (K 0321), (courtesy \"{O}sterreichische Nationalbibliothek).
     The corresponding text is featured below.
     \\
     \\
\textsf{\textbf{Coptic:}}~\begin{coptic}h1wcte ened1tefythn mpbol ntefri nyomnt nh1oou ntetmlaau taioc efitc\end{coptic}\\

\textsf{\textbf{Reference:}} ~\textit{such that if he throws his tunic out of his cell for three days no one will pick it up to wear it}\\

\textsf{\textbf{GPT-4.1 Translation:}}~\textit{So that his light would not shine before people. You should not do this at all.}\label{ex:bad-translation}\end{mybox} 
          \caption{Reference translation and the baseline translation for Coptic text, (corresponds to the excerpt at the top). Even large models such as GPT-4.1 provide fluent yet fundamentally incorrect translation without augmentation.}
    \label{fig:top-right-box}
\end{figure}

However, prompt augmentation using in-context learning (ICL), by integrating bilingual glosses for vocabulary items~\cite{ghazvininejad-et-al-2023-dictionary}, has offered a promising direction. This is notably useful to overcome the limitations in languages where we have no hope of obtaining sufficient amounts of raw text data. We can leverage LLMs' fluency, especially when there is a dictionary or glossary for the source language, and the target language is high-resource, i.e. for translation from the LRL to the HRL. At the same time, simple glosses are inherently limited to single, or in some cases few-word listed expressions, and cannot inform models about the grammar and specific constructions of the LRL, leaving models to generate target sentences based on plausible configurations of target lexical item senses.

\paragraph{} In this paper, we target \textbf{Coptic}~(specifically the Sahidic dialect)\footnote{ISO 639-2 code: 'cop', Glottolog: \url{https://glottolog.org/resource/languoid/id/copt1239}. Sahidic: \url{https://glottolog.org/resource/languoid/id/sahi1241}}, a low-resource language\footnote{It is considered an endangered language with no L1 speakers~\cite{ethnologue_2026}.} for which LLMs have little or no coverage out of the box, and in which grammatical constructions often convey differences in meaning that are not obvious from content words alone.

Coptic, illustrated in \cref{fig:top-right-box}, is the last phase of the indigenous language of Egypt, spoken and written primarily in the first millennium CE. It forms part of the Egyptian branch of the Afro-Asiatic language family. It is an agglutinative, head initial language with a complex system of auxiliaries, grammatical gender and number, aspect and tense distinctions. As the language of Christian Egypt in the Hellenistic period, it is crucial for our understanding of the history of religion and of the Mediterranean in Late Antiquity, and it remains in use today as the heritage language of Egyptian Christian Copts in Egypt and the diaspora. 

However, data for Coptic is scarce, as are experts in the language, leading to many cataloged manuscripts in museum libraries remaining undigitized, or digitized but untranslated. This motivates our work to strive for high quality machine translation outputs, which, even if not sufficiently accurate to be used as-is, could reduce the effort for experts to correct. 

Testing current LLMs on out-of-the-box translation quality for Coptic quickly reveals their inadequacy. For example, the sentence taken from the manuscript in  \cref{fig:top-right-box} refers to an ascetic monk's rags being so tattered that, if left outside unattended for three days, no one would bother to steal them. Even very large models, such as GPT-4.1, produce a fundamentally incorrect translation, indicating the model's inability to process the language as-is.

For languages such as Coptic, exploring translation based on the methodologies and resources that are specifically available is important~\cite{bapna2022building}. Although dictionary-based prompt augmentation already improves machine translation for LRLs, and has even recently been applied to Coptic \cite{miyagawa-2025-rag}, the inability to encode grammatical relations by simply listing lexical items creates a ceiling on translation quality, which this work aims to address. 

In particular, this paper is the first to explore whether the addition of syntactic information, such as \text{Universal Dependencies} or UD~\cite{de-marneffe-etal-2021-universal}, can augment in-context MT for LRLs such as Coptic when provided with lexical information. We compare various strategies: baseline translation with no augmentation, with dictionary augmentation, with syntactic augmentation, and with both dictionary and syntactic information. We test both open-weight and closed-source models (\cref{sec:incontext-coptic-translation}). 

We find that while lexical information remains more important than syntactic information, adding syntactic information based on UD parses significantly improves performance over lexicon integration alone across all model sizes (\cref{sec:results}). We additionally provide qualitative and quantitative analyses examining the effects of different operationalizations, the impact of automatic parses (compared to gold parses), and the differences between biblical and non-biblical samples (\cref{sec:analysis}). All code is made available on the GitHub repository for this paper.\footnote{The code is available at \url{https://github.com/gucorpling/in-context-coptic-translation}}

\section{Background}
\label{sec:background}

LLMs have become the standard for MT in many settings~\cite{kocmi-etal-2025-findings}. However, methods and models vary based on the languages, resources, and domains. This paper specifically focuses on using LLMs in an ICL setting, utilizing lexicon and syntactic analysis. Supervised training methods have shown success when parallel training data can be collected or built, but this does not extend to LRLs such as Coptic. In these scenarios, both expert and non-parallel data have been used for augmentation to support low-resource MT.

\paragraph{Lexicon}
Bilingual lexicons are a common resource used for low-resource MT, including in LLM-based approaches, especially for translating from a LRL to HRL. \citet{ghazvininejad-et-al-2023-dictionary} showed how simple textual demonstrations can be added to prompts to improve performance. 
\citet{lu-etal-2024-chain} showed how a chain of bilingual lexicons can be used to provide a bridge between source and target languages. A range of parameters can influence ICL formulations, especially for LRLs~\cite{court-elsner-2024-shortcomings}, partly due to the context-use challenges of LLMs~\cite{liu-etal-2024-lost}. 

\paragraph{Grammar}\label{sec:background-grammar} ICL MT provides a unique opportunity to provide information about the grammar of a language to be used in-context to help with translation. This has been studied in various settings, such as retrieval from a grammar with extraction of excerpts~\cite{tanzer-2024-mtob} or constructions of an expert set of grammar rules~\cite{zhang-etal-2025-read}. \citet{pei-etal-2025-understanding} investigated ICL MT for the LRL of Manchu (\texttt{mnc}), using dictionary entries (with morphological analysis), parallel examples, and extracted grammar excerpts. They found grammar excerpts to be less useful in the presence of the other two resources. Our paper utilizes dictionary entries and morphological analysis (although not as high-level components). It additionally focuses on using grammatical information derived from the input, rather than relying on the explicit grammar of language itself as in~\citet{pei-etal-2025-understanding}.

\paragraph{Grammatical information}
Unlike grammar books and excerpts, grammatical information is relatively less explored as a resource for augmentation for ICL. Such information can be broad, from morphological analysis~\cite{elsner-needle-2023-translating, pei-etal-2025-understanding} to annotated linguistic data such as UD. Additional challenges are the variations in the frameworks and the nature of the data (such as genre) annotated. The heart of our work explores this scenario, specifically with syntactic information, using Universal Dependencies.

\paragraph{Universal Dependencies treebanks for MT}
UD treebanks provide annotations for over 150 languages using a consistent grammatical framework. While UD treebanks are used predominantly for the study of language and the improvement of monolingual and multilingual parsing, they have also been used in conjunction with other tasks, including MT. For example, \citet{nagy-etal-2023-treeswap} used UD trees as a substrate to create additional parallel data for MT. To our knowledge, our paper is the first to utilize UD as a source of grammatical information for MT.

\paragraph{Coptic MT}
Work targeting Coptic in particular has only recently emerged, with papers using English \cite{wannaz-miyagawa-2024-assessing,saeed-etal-2024-nile,miyagawa-2025-rag}, French \cite{chaoui2025neuralmachinetranslationcopticfrench} and Arabic \cite{saeed-etal-2024-nile} as HRL targets. The source data in all of these papers is in Sahidic Coptic, the classical dialect of the language which we also target, except for \citet{saeed-etal-2024-nile}, who use the later attested Bohairic dialect of Coptic. Although these studies have shown that fine-tuning models can somewhat improve results \cite{chaoui2025neuralmachinetranslationcopticfrench}, translation outputs are still far from ready for public use, and therefore do not yet satisfy our aspirations to make more Coptic texts available to the public via translation.

\section{In-Context Coptic Translation}
\label{sec:incontext-coptic-translation}
In-context translation approaches for LRLs, such as Coptic, augment input with examples and other forms of language or linguistic data, rather than relying solely on supervised training data. These methods focus on selecting and presenting auxiliary information, such as lexical or syntactic details, that can guide and improve translation behavior.

Within this framework, translation prompts are explicitly constructed to incorporate relevant resource-derived information alongside the source sentence. This design allows linguistic knowledge to be selectively included and evaluated, providing a flexible means of improving translation quality in low-resource settings. For our translation task, Coptic is the source language, English is the target language, and it is a segment-level translation. This is mainly a function of the resources ~(discussed further in \cref{sec:resources}), but it also allows us to focus on both designing expert based methods (\con~in \cref{sec:components}), and providing more detailed analyses (\cref{sec:analysis}).

\label{sec:LLMs}
We experiment with several open models, but we will focus on Gemma models for our analysis, which showed the best performance during development. We also use GPT-4.1 as the closed reference models (see \cref{tab:models} for the full list).

\begin{table}[t]
    \centering
    \begin{tabular}{ll}
         \textbf{Model (Size)} & \textbf{Technical Report}\\
         \toprule
         % \textbf{Main models} & \\
         Gemma3(-it) (12B) &  \citet{Gemma3technicalreport}\\
         Gemma3(-it) (27B) &  \citet{Gemma3technicalreport}\\
         
         GPT4.1 (NA) & \citet{gpt4technicalreport}\\
         \midrule
         % \textbf{Other models} & \\
         GPT4.1 Mini (NA) & \citet{gpt4technicalreport}\\
          Llama3.1(-Inst) (8B) & \citet{grattafiori2024llama3herdmodels}\\
         Aya-Expanse (8B) & \citet{2024ayamodelinstructionfinetuned}\\
         Aya-Expanse (32B) & \citet{2024ayamodelinstructionfinetuned}\\
         \bottomrule
    \end{tabular}
    \caption{Models used for our experiments with \dev~data. We report results for \test~and ostraca from \citet{wannaz-miyagawa-2024-assessing} only for the first three models. We do not use the base variant of models, so will not explicitly refer to those with the (-it) suffix for the rest of the paper.}
    \label{tab:models}
\vspace{-10pt}
\end{table}

\subsection{Data}
\label{sec:data}
The largest publicly available dataset of Coptic sentences and translations comes from Coptic Scriptorium \cite{SchroederZeldes2015}. This includes $2.3$M tokens of Coptic, of which $\sim$1.43M are in the Sahidic dialect, though translations exist for only $1.21$M tokens of this data, including most books of the Old and New Testament. This is also the dataset used for translation by \citet{chaoui2025neuralmachinetranslationcopticfrench}, who selected four books of the Bible (1 Corinthians, Mark, Galatians, and Hebrews) as their test set. Since we already have translations of the Bible and suspect that LLMs can more easily produce translations of Bible verses provided they can identify the target verse (for example, through the occurrence of proper names), in this paper we additionally analyze MT results between Biblical and non-Biblical texts.

\paragraph{UD Treebank: Translation dataset and parses} Since our method is the first to use UD parses for MT, we require data for which gold standard syntax trees exist, which allows us to assess the impact of cascading parser errors by comparing predicted and gold parse inputs. We therefore focus on the manually treebanked subset of Coptic Scriptorium available in the Sahidic UD Coptic treebank \cite{zeldes-abrams-2018-coptic}. This contains over $\sim$60K tokens (2,387 sentences) with translations and covers a range of genres including Bible translations, indigenous Coptic hagiography, sermons, and documentary materials. We use the UD treebank's standard splits of \dev~(380 sentences, of which 182 are from the Bible) and \test~(405 sentences) as our core data. 

\paragraph{Ostraca: Out-of-domain translation dataset}
\label{sec:data-ostraca}
We also report results on out-of domain data from \citet{wannaz-miyagawa-2024-assessing}, which contains 4 ostraca (21 sentences), with previous MT results.\footnote{
We do not compare results with \citet{saeed-etal-2024-nile} since their data targets Bohairic (\url{https://glottolog.org/resource/languoid/id/boha1242}), a substantially different dialect.}

\subsection{Resources}
\label{sec:resources}
Broadly, we utilize two sets of information: a fixed dictionary and a sample-specific syntactic analysis NLP pipeline.

\paragraph{Dictionary} The Coptic Dictionary~(hereafter the dictionary)\footnote{\url{https://coptic-dictionary.org/}} from \citet{feder-etal-2018-linked} and updated with the lemma list from \citet{burns2020coptic_lexicon} acts as the as our sole dictionary resource. 
The dictionary was constructed by integrating multiple bilingual lexicons, providing extensive and elaborate information for lexical items.
The dictionary covers over 10K entries, but multiple entries can exist for a single surface form.

\paragraph{Syntactic analysis}
\label{sec:coptic-nlp}
To add grammatical information to our input sample, we utilize Coptic-NLP~\cite{zeldes-schroeder-2016-nlp}, which performs automatic segmentation of agglutinative Coptic word forms into tokens and produces full UD parses. These parses include both Google Universal POS tags \cite[][upos]{PetrovDasMcDonald2012}, language-specific tags \cite[][Coptic xpos tags]{ZeldesSchroeder2015}, morphological features, and language-of-origin for each word (for example, identifying Greek loan words). We use this information in multiple ways below, including raw parser output, templated verbalization into English (e.g.~`The subject of the verb X is the noun Y'), and verbalized translation instructions for special constructions (see below).  

\subsection{Components}
\label{sec:components}

To make the best use of these resources in the ICL setting, we design specific `components' that handle processing and representing information from the resources. We define four components: \lexicon, \dep, \con, and \conll~. Each draws on the resources and provides complementary information about the source sentence.

\paragraph{Lexicon \texttt{LEX}}

Our \lexicon~component is designed to map the structured information from the dictionary to the input context. We use syntactic analysis for the input sentence to inform this. We use POS and morphological information (lemma and segmentation) to search the lexicon for dialect specific translations. To control prompt length and relevance, we further filter retrieved entries, retaining the most relevant entry- and sense-level hierarchical information for inclusion in the instruction context. Further details about the \lexicon~component are provided in \cref{sec:appendix-lexgridsearch}.

\paragraph{Syntax - Dependency \dep}
Our \dep~component verbalizes the syntactic structure from the parses for inclusion into the instruction. For each input sentence, we extract the head–dependent relations and verbalize them as short, plain English statements (e.g. `\begin{coptic}n\end{coptic} is the case marking of \begin{coptic}yoyou\end{coptic}'). This representation provides explicit syntactic relations between tokens, which are not necessarily inferrable from surface word order alone. Multiple parameters control the granularity and content of the syntactic information added to the instruction, such as the selected UD label set, selected parts of speech (POS), and disambiguation for repeated tokens. In all configurations, the dependency descriptions are rendered in plain English and the dependency section as a whole can be added to the instruction similarly to the lexicon component. Further details about the \dep~component are provided in \cref{sec:appendix-depgridsearch}.

\paragraph{Syntax - Construction \con}
\con~ is the most bespoke component in the paper. We perform a manual analysis of model errors on the development set\footnote{One of the authors has training in Coptic.}, and in conjunction with the syntactic analysis, identify specific grammatical constructions for which the pilot model demonstrates error.\footnote{We used results from GPT4.1 mini for this analysis.} In particular, we identify $26$ tree configurations and formalize their characteristic dependency subgraphs using \texttt{DepEdit}~\cite{peng-zeldes-2018-roads}, a Python templating library for UD trees. These constructions range from simple and general to highly specific. 

For example, imperatives are trivially recognizable using UD morphological features. Like English, Coptic verbs with no subject have an imperative meaning, but merely glossing words for translation can easily miss the need for an imperative translation, which a targeted instruction can clarify. On the complex end, one construction involving postponed subjects (roughly, examples like `he went to the desert, that is, the monk') triggers an explanation of how Coptic uses this configuration.

This module also uses a dedicated subroutine to transliterate any words tagged as proper nouns (\texttt{PROPN}) into Latin characters, creating a sub-instruction which states that the word is a name, accompanied by the transliteration. For more detailed examples see \cref{sec:appendix-constructions}.

\paragraph{Syntax -~\conll}
\label{sec:raw_conllu}

Since we use UD and Coptic NLP for syntactic analysis, a low effort option is to effectively dump the raw output (CoNLL-U format) in the instruction. Given the prevalence of the CoNLL formats and similar structured formats on the internet, it is necessary to explore this low-effort representation operationalization.\footnote{An example of CoNLL-U is given in the appendix \cref{fig:conllu}.}

\subsection{Instruction Design}
\label{sec:instructon-design}

\begin{figure}[th]
    \centering
    \tcbset{colback=white}
    \begin{mybox}\sf
        You are a professional Coptic-to-English translator tasked with providing translations suitable for
use in United States (en\_US). $\cdots$ Please translate
the following Coptic text into English (en\_US): $\cdots$\\
(From \lexicon) For the translation task, you are given dictionary entries for Coptic. $\cdots$\\
(From \conll) The raw conllu data for the sentence is in the CONLL-U format:\\
1  \begin{coptic}m\end{coptic} \begin{coptic}n\end{coptic}  ADP  PREP \_  3  case  \_  \_$\cdots$\\
(From \dep) The dependency information for the sentence is: \begin{coptic}rmh\end{coptic} is the root. $\cdots$\\
(From \con) The information about specific constructions $\cdots$ The dislocated element \begin{coptic}pai\end{coptic} is
a repeated reference to the pronoun dependent of the predicate \begin{coptic};w.\end{coptic}
Using all the information provided above, now please translate the sentence into English(en\_US).
Remember your source sentence is: \{\textit{source}\} .The English translation is:

    \end{mybox}
    \caption{A condensed example of how the different information is added to the instruction. Information added from each component is based on the experimental setting (\cref{text:settings}). \all~would include information from all components. More details of different parts are provided in \cref{sec:appendix-instruction-design}.}
    \label{fig:instruction-example}
    \vspace{-10pt}
\end{figure}
Our instruction (or prompt) consists of a base instruction with additional information derived from our components, and closed out with consistency cues (see \cref{fig:instruction-example}). We adapt the base instruction from \citet{wmt25-instructions}. For dictionary-based information, we adapt the instructional framework from \citet{pei-etal-2025-understanding}. We add information from each of the components for the setting (\cref{sec:settings}) with a small textual header indicating the section. We additionally added some consistency cues similar to \citet{pei-etal-2025-understanding} to help improve model responsivity.

\subsection{Metrics}
\label{sec:metrics}
MT has a wide variety of metrics targeting different aspects of evaluation~\cite{lavie-etal-2025-findings}. We use BERTScore~\cite{bertscore}, specifically the Avg. F1 of BERTScore, as our primary metric for our development work and analyses in this paper. It was chosen based on the previously seen correlations with human evaluation, even in comparison to LLM-as-judge settings for MT~\cite{lavie-etal-2025-findings}.

We additionally report BLEU~\cite{papineni-etal-2002-bleu} for comparison to \citet{wannaz-miyagawa-2024-assessing} on the ostraca. For more effective comparison with their reported results, we further include METEOR~\cite{banerjee-lavie-2005-meteor} for this dataset. 

Additional discussion and details about the metrics for completeness and reproducibility are provided in \cref{sec:appendix-metrics}

\subsection{Settings}
\label{sec:settings}

We evaluate a range of prompting configurations that incrementally augment the baseline translation setting with additional components. Our baseline setting consists of a simple translation instruction without any auxiliary information. We then run experiments that add individual linguistic components on their own, as well as combinations of multiple components. \label{text:settings} 

We consider settings with a single component \lexicon, \conll, \con, or \dep. We additionally examine two combination settings, \dep+\con~and \lexicon+\syntax~(which is effectively \lexicon+\conll+\con+\dep). The \dep+\con~setting groups the designed syntactic components but not the raw parse component \conll~to isolate the effect of this distinction.

\paragraph{\lexicon~and~\dep~grid search}
Both \lexicon~and \dep~have parameters that determine the kind and amount of information incorporated into the instruction context.
We perform a targeted grid search to determine our best lexicon and dependency parameter values using our pilot model (GPT-4.1 mini) and a diagnostic subset of \dev~data consisting of 20 samples drawn from an initial baseline run. We used 10 sentences with the highest translation quality scores (the 10 easiest) and the 10 with the lowest scores (the 10 hardest). We then shortlisted four parameter configurations and choose the final configuration based on the \dev~split using BERTScore F1 (see \cref{sec:metrics}) as the primary criterion. Further information on this is in \cref{sec:appendix-sys-and-experiments}. 

\paragraph{With gold parses}
\label{sec:with-gold}

To analyze the use of automatic parses compared to gold-standard annotations, we additionally conduct experiments only on the \dev~split using gold-standard UD parses in place of automatically generated parses. We conduct and report this experiment only for the Gemma models.

\begin{table*}[!ht]
    \centering
    \resizebox{\textwidth}{!}{
\begin{tabular}{lrrrrrr} 
             & \multicolumn{2}{c}{\textbf{Gemma-12b}}& \multicolumn{2}{c}{\textbf{Gemma-27b}} &\multicolumn{2}{c}{\textbf{GPT-4.1}} \\
\toprule
\textbf{Setting} & \textbf{BLEU} ($\Delta$) & \textbf{BertScore} ($\Delta$)  & \textbf{BLEU} ($\Delta$) & \textbf{BertScore} ($\Delta$) & \textbf{BLEU} ($\Delta$) & \textbf{BertScore} ($\Delta$)  \\
\midrule
\multicolumn{7}{c}{\textbf{\test~split}}\\
\midrule
Baseline & 60.65 (0.00) & 0.8363 (0.0000)  & 19.18 (0.00)& 0.8385 (0.0000) & 13.56 (0.00)& 0.9012 (0.0000)\\[3pt]
\underline{\lexicon}  & 36.84 (-23.81) & 0.8551 (0.0187) & 9.41 (-9.77) & 0.8565 (0.0181) & 13.56 (0.00) & 0.9152 (0.0140) \\
\conll  & 13.56 (-47.09) & 0.8489 (0.0126) & 13.56 (-5.62) & 0.8547 (0.0162) & 13.56 (0.00) & 0.9056 (0.0044) \\
\con  & 5.50 (-55.15) & 0.8511 (0.0148) & 19.18 (0.00) & 0.8518 (0.0133) & 13.56 (0.00) & 0.8998 (-0.0014) \\
\dep  & 60.65 (0.00) & 0.8420 (0.0057) & 13.56 (-5.62) & 0.8417 (0.0033) & 13.56 (0.00) & 0.9014 (0.0002) \\
\dep+\con & 5.50 (-55.15) & 0.8502 (0.0139) & 13.56 (-5.62) & 0.8530 (0.0146) & 13.56 (0.00) & 0.9030 (0.0018) \\
\textbf{\all} & 13.56 (-47.09) & 0.8707 (0.0344) & 9.41 (-9.77) & 0.8746 (0.0361) & 13.56 (0.00) & 0.9195 (0.0183) \\
    \bottomrule
    \end{tabular}
    }
    
        \resizebox{\textwidth}{!}{
\begin{tabular}{lrrrrrr} 
    \multicolumn{7}{c}{\textbf{\dev~split\hphantom{t}}}\\
    \midrule
    Baseline & 16.45 (0.00) & 0.8342 (0.0000) & 10.30 (0.00)& 0.8375 (0.0000)& 14.58 (0.00)& 0.9015 (0.0000) \\[3pt]
    \underline{\lexicon} & 5.46 ($-$10.99)& 0.8593 (0.0250)& 14.46 (4.16)& 0.8611 (0.0235)& 10.72 ($-$3.86)& 0.9139 (0.0124)\\
    \conll & 27.49 (11.04)& 0.8453 (0.0111)& 14.86 (4.56)& 0.8522 (0.0147)& 28.21 (13.62)& 0.9048 (0.0033)\\
    \con & 20.26 (3.81)& 0.8499 (0.0157)& 15.43 (5.14)& 0.8509 (0.0133)& 12.77 ($-$1.81)& 0.9000 ($-$0.0016)\\
    \dep & 28.66 (12.21)& 0.8405 (0.0063)& 15.28 (4.98)& 0.8414 (0.0039)& 16.30 (1.72)& 0.9020 (0.0004)\\
    \dep+\con& 21.65 (5.20)& 0.8490 (0.0147)& 10.38 (0.09)& 0.8509 (0.0134)& 12.79 ($-$1.79)& 0.9018 (0.0003)\\
    \textbf{\all} & 7.71 ($-$8.74)& {0.8722 (0.0380)}& 18.84 (8.54)& {0.8736 (0.0360)}& 12.69 ($-$1.89)& {0.9169 (0.0154)}\\
    \bottomrule
    \end{tabular}
    }
    \caption{Results for the \test~split and \dev~split  Gemma and GPT-4.1 model across different settings (differences in performance are significant for $p < 0.0001$.). Syntactic information provides complementary benefits beyond what we see from just lexicon. The \all~setting (bolded) combining \lexicon, \conll, and \dep+\con~performs the best for all models. The \lexicon~setting, known to be effective, is underlined for ease of comparison. BLEU is reported for completeness (see \cref{sec:metrics}).}
    \label{tab:test-components}\label{tab:dev-components}
\end{table*}

\section{Results}
\label{sec:results}

Our strategy of adding syntactic information leads to performance improvements in both open and closed models, and across \test~(shown in \cref{tab:test-components})\footnote{BLEU is also reported for consistency with a relaxed setting on \test. It is unstable with the \test~data due to variation in LLM translations and drastic LLM errors, making it unreliable (more details in \cref{sec:bleu-test-unreliable}). For \dev~and ostraca it is reported with default signature.} and ostraca (shown in \cref{tab:ostraca-result}).

Across the three models, adding the \lexicon~component yields statistically significant improvements over the baseline (for all of \dev, \test, and ostraca) as expected. Likewise, adding \syntax~also results in significant gains in BERTScore (see \cref{sec:appendix-significance} for significance test details). This indicates that syntactic information provides complementary benefits (see \cref{tab:test-components})
beyond what we see from just lexicon. These trends are consistent across both Gemma models and GPT-4.1.

The highest augmentation setting \all~shows the best performance across all our models. This suggests that combining both lexical and syntactic information provides cumulative improvements to the model. We disambiguate the contribution of each syntactic component when added on top of \lexicon~in the next section~( \cref{sec:lex-ablations}).

The same trends are seen in \dev~(see \cref{tab:dev-components}) and across other models. However, this was not consistent for all metrics. 

\begin{table}[!ht]
    \centering
    \begin{tabular}{lrrr}
    \textbf{Model} & \textbf{BLEU} & \textbf{BertScore} & \textbf{MET.} \\
        \midrule
        \multicolumn{4}{l}{\textbf{From this paper}}\\[3pt]
        Gemma & & & \\
        \hphantom{G}12b \all & 7.82 & \underline{0.8850} & 0.32 \\
        \hphantom{G}27b \all & \underline{16.19} & 0.8781 & 0.34 \\[3pt]
        GPT-4.1 & & & \\
        \hphantom{G}\all & 17.99 & \underline{0.9046} & \underline{0.53} \\
        \hphantom{G}\dep & \underline{18.15} & 0.8859 & 0.38 \\[3pt]
        \multicolumn{4}{l}{\textbf{From \citet{wannaz-miyagawa-2024-assessing}}}\\[3pt]
        Claude Opus & 20.02 & - & 0.46\\
       Claude Haiku & 11.52 & - & 0.35\\
        CopticTrans & 8.43 & - & 0.30\\
        \bottomrule
        \end{tabular}
    \caption{The best Gemma and GPT-4.1 settings for ostraca~(\cref{sec:data-ostraca}).
    The BLEU and METEOR scores for Claude Opus and Haiku~\cite{Claude3S}, and CopticTranslator are from \citet{wannaz-miyagawa-2024-assessing}. Our open models are comparable, and GPT-4.1 exceeds previous model on METEOR but not BLEU. Full results are in the appendix (\cref{tab:ostraca-result}).}
    \label{tab:ostraca-highlight}
    \vspace{-15pt}
\end{table}

For the ostraca, our best open-models results are comparable to the closed models reported by \citet{wannaz-miyagawa-2024-assessing}, and GPT-4.1 versions perform better than any reported system on METEOR but not BLEU (see \cref{tab:ostraca-highlight}).
The \all~setting, which was shown to be the most useful in \dev~and \test, also has the best performance on this set for both open models and closed models.\footnote{Both papers used the default settings from \texttt{sacrebleu} python package for BLEU.}

\section{Analysis and Findings}
\label{sec:analysis}
\begin{table*}[!ht]
\resizebox{\textwidth}{!}{%
\begin{tabular}{llr}
 \textbf{Setting} & \textbf{Translation}    & \textbf{$\Delta$ F1} \\
 \midrule
 \textbf{Gemma-12b }& & \\
 Reference        & \textit{He also said that Abba Pambo said:``This is the way that it is   fitting for the monk to wear his clothe''} & --\hphantom{.0000}             \\
Baseline         & \textit{And with all the powers through Jesus Christ our Lord, to give life to   those who are sick.}               & 0\hphantom{.0000}              \\
\lexicon          & \textit{Again, they say that the abbot has garments that it is fitting to bear,   indeed, for the monks.}           & +0.0790          \\
\all          & \textit{Again, he said that the abbot Pambo said that he is fitting for the monks   to bear the garments.}          & +0.1004         \\[3pt]
\cmidrule(lr){1-3}
Reference        & \textit{Why do we now turn around and entangle us in them?}                                                         & --\hphantom{.0000}             \\
Baseline         & \textit{And on the flesh of his body.}                                                                              & 0\hphantom{.0000}              \\
\lexicon          & \textit{Again, let it turn around, and let it be wrapped in the upper part.}                                        & +0.0220          \\
\all          & \textit{Why do we turn around to embrace the upper parts of them?}                                                  & +0.0912         \\
\midrule
\textbf{Gemma-27b} & & \\
Reference        & \textit{Herodias set herself against him, and desired to kill him, but she couldn't}                              & \textbf{--\hphantom{.0000}}    \\
Baseline         & \textit{The knowledge of God is the beginning of life. {[}…{]}}       & 0\hphantom{.0000}              \\
\lexicon          & \textit{He is angry with the anger of the people. He wishes to kill with his   strength}                            & +0.0190          \\
 \all          & \textit{Herodias is angry, she wishes to kill him, but she is not able.}                                            & +0.0981         \\[3pt]
 \cmidrule(lr){1-3}
Reference        & \textit{he gave them to the poor}                                                                                   & --\hphantom{.0000}             \\
 Baseline         & \textit{Into the water}                                                                                             & 0\hphantom{.0000}              \\
 \lexicon          & \textit{Give to the poor}                                       & +0.0392         \\
\all          & \textit{He gave to the poor}                                                                                        & +0.0898         \\
\midrule
\textbf{GPT-4.1}  & &\\
Reference        & \textit{And the mindless, whose acts of stupidity   abound, are like bats.}                                         & \textbf{--\hphantom{.0000}}    \\
Baseline         & \textit{And they will weep bitterly in great lamentation; they will stay up   through the nights.}                  & 0\hphantom{.0000}              \\
\lexicon          & \textit{And hearts that are greatly exalted among measures of grain; they are   likened to stars.}                  & +0.0079         \\
\all          & \textit{And the mindless ones, who are exceedingly great in foolishness, are like   bats.}                          & +0.0776         \\[3pt]
\cmidrule(lr){1-3}
Reference        & \textit{Now on the mountainside there was a great herd of pigs feeding.}                                            & --\hphantom{.0000}             \\
 Baseline         & \textit{But she did not die, for a great multitude of angels appeared to her.}                                      & 0\hphantom{.0000}              \\
 \lexicon          & \textit{But there was no place there on the mountain for the great herd of swine   to feed.}                        & +0.0553         \\
\all          & \textit{But there was a place near him on the mountain, where a large herd of   pigs was feeding.}                  & +0.0753         \\
\bottomrule
\end{tabular}
}
\caption{Sample outputs with no augmentation (Baseline), with lexicon information added (\lexicon), and \all~(also includes all syntax components), as well as BERT-Score F1 improvement $\Delta$ for Gemma and GPT-4.1 models.}
\label{tab:err-ana}
\vspace{-5pt}
\end{table*}

To better understand the impact of different components, and nature of the data, we perform additional analyses on \dev. 

\paragraph{Error analysis} Table \ref{tab:err-ana} provides some examples for qualitative error analysis, which illustrate how models leverage \lexicon~and \syntax. In the baseline setting, open models can generate totally unrelated translations, since they have almost no support for Coptic. Lexicon augmentation induces relevant vocabulary in outputs yielding lexically accurate outputs even in the 12b model, and near-perfect lexical choices in translations from the 27b model (e.g., `Give to the poor' and `wishes to kill'). Adding syntax helps both in connecting the subject and verb for `He gave to the poor' and in complex nesting for `he said that the abbot Pambo said', a relatively unlikely nested reported speech, which the lexicon-only augmentation fails to capture. 

Effectiveness of transliteration in~\con~for nouns tagged \texttt{PROPN} is also apparent, leading to correct renditions of `Herodias' and `Pambo'. GPT-4.1 follows a similar pattern: baseline translations seem almost random, and \lexicon~augmentation supplies some relevant words; however \syntax~augmentation allows for recognition of relations such as relative clauses (`mindless ones who are...') and disambiguating senses -- the negation in `there was no place' is likely due to the Coptic negation `n' being a homonym with the preposition `of', a distinction the UD tree makes clear by explicitly identifying prepositions.

\paragraph{Differing forms of syntax}

The improvements from syntactic information varies across models and depends on how it is incorporated. We evaluate four syntax-only settings: \conll, \con, \dep, and \dep+\con. When added to our baseline, as shown in \cref{tab:dev-components}, GPT-4.1 and Gemma-27b achieve their largest F1 improvements even with just the \conll~component. In contrast,  Gemma-12b improves the most from \con, actually seeing the least improvement from the \conll~component.

\begin{table}[!htb]
\label{sec:lex-ablations}
    \centering
    \begin{tabular}{lrrr} 
     & \multicolumn{2}{c}{\textbf{Gemma}} & {\textbf{GPT-4.1}}\\
     \midrule
        &  \multicolumn{1}{c}{\textbf{12b}}&  \multicolumn{1}{c}{\textbf{27b}} &  \\
        \midrule
        \textbf{Setting}& \multicolumn{3}{c}{\textbf{BertScore}}\\
        \midrule
    \lexicon & 0.8593 & 0.8611 & 0.9139\\[3pt]
    +\conll & +0.0053 & +0.0076& \underline{\textbf{+0.0042}}\\
    +\con & \underline{+0.0107}& \underline{+0.0094}& $-$0.005\\
    +\dep & +0.0005 & $-$0.0009& +0.0001\\[3pt]
    % +\dep & & &\\
    % \hphantom{+} +\conll & +0.0088&  +0.0121& +0.0025\\
    % \hphantom{+} +\con & +0.0147& +0.0150& $-$0.0001\\
    +\dep +\conll & +0.0053&  +0.0076& +0.0036\\
    +\dep +\con & +0.0112& +0.0105& $-$0.0010\\[3pt]
    % .8722 - .8593, .8736 - .8611, 0.9169 - 9.9139
        \lexicon+\syntax & \textbf{+0.0129} & \textbf{+0.0125} & +0.0030 \\
    \bottomrule
    \end{tabular}
    \caption{\lexicon~+ ablation results on \dev. Each row displays the change from the baseline of \lexicon. We underline the best single-component score and bold the best overall score. Syntactic augmentation consistently improves performance, with strong score improvements from \con, \conll, and \syntax.} 
    \label{tab:dev-lex-ablation}
    \vspace{-10pt}
\end{table}

When added on top of \lexicon, as seen in \cref{tab:dev-lex-ablation}, all syntactic components yield gains for the Gemma models. \con~and \conll~provide larger improvements than \dep~individually. In GPT-4.1, the simple use of \conll~outshines other syntactic components and this model sees limited or even negative gains from other components.

Improvements from \conll~alone show the potential for a low-effort way of improving performance, given high-quality NLP tools for the language. 

We generally see that while syntax alone is not as useful, it can provide significant complementary gains when layered on top of \lexicon.

\begin{table}[!htb]
    \centering
    \begin{tabular}{lrrrr}
         & \multicolumn{2}{c}{\textbf{Gemma-12b}} & \multicolumn{2}{c}{\textbf{Gemma-27b}} \\
         \midrule
         &\multicolumn{1}{c}{\textbf{Auto}}&\multicolumn{1}{c}{\textbf{Gold}} & {\textbf{Auto}}&\multicolumn{1}{c}{\textbf{Gold}}\\
        \midrule
        \textbf{Setting} & \multicolumn{4}{c}{\textbf{BertScore}}\\
        \midrule
        
        Baseline & 0.8342 & -  & 0.8375 &-  \\
        \conll & 0.8453 & {0.8499}  & 0.8522 &{0.8545}  \\
        \con & 0.8499 & 0.8502  & 0.8509 &0.8512  \\
        \dep & 0.8405 & 0.8411  & 0.8414 &0.8413  \\
         \dep+\con & 0.8490 & 0.8483  & 0.8509 &0.8518  \\
         \bottomrule
    \end{tabular}
   
    \caption{The performance of different syntax related settings with the automatic parses (all other results), and explicit use of gold parses. The different in performance between the two settings exists, but is not drastic. Results of all the different settings, with multiple metrics can be seen in appendix (\cref{tab:dev-gold}).}
    \label{tab:dev-gold-highlights}
    \vspace{-10pt}
\end{table}

% \begin{table*}[!htb]
%     \centering
%     \resizebox{\textwidth}{!}{
% \begin{tabular}{lrrrrrrr} 
%         & \multicolumn{2}{c}{\textbf{Gemma-12b}}& \multicolumn{2}{c}{\textbf{Gemma-27b}} &\multicolumn{2}{c}{\textbf{GPT-4.1}} \\
%         \toprule
%     \textbf{Setting} &  \textbf{BLEU} ($\Delta$)  & \textbf{BertScore} ($\Delta$)  &  \textbf{BLEU} ($\Delta$)  & \textbf{BertScore} ($\Delta$)  &  \textbf{BLEU} ($\Delta$) & \textbf{BertScore} ($\Delta$)\\
%     \midrule
%     Baseline & 16.45 (.00) & .8342 (.0000) & 14.86 (.00)& .8375 (.0000)& 14.58 (.00)& .9015 (.0000) \\[3pt]
%     \all & 7.71 ($-$8.74)& \textbf{.8722 (.0380)}& 15.28 (.42)& \textbf{.8736 (.0360)}& 12.69 ($-$1.89)& \textbf{.9169 (.0154)}\\
%     \conll & 27.49 (11.04)& .8453 (.0111)& 10.30 ($-$4.56)& .8522 (.0147)& 28.21 (13.62)& .9048 (.0033)\\
%     \con & 20.26 (3.81)& .8499 (.0157)& 15.43 (.57)& .8509 (.0133)& 12.77 ($-$1.81)& .9000 ($-$.0016)\\
%     \dep & 28.66 (12.21)& .8405 (.0063)& 14.46 ($-$.40)& .8414 (.0039)& 16.30 (1.72)& .9020 (.0004)\\
%     \lexicon & 5.46 ($-$10.99)& .8593 (.0250)& 18.84 (3.98)& .8611 (.0235)& 10.72 ($-$3.86)& .9139 (.0124)\\
%     \dep+\con & 20.94 (13.23)& .8490 (.0147)& 10.38 ($-$4.48)& .8509 (.0134)& 12.79 ($-$1.79)& .9018 (.0003)\\
%     \bottomrule
%     \end{tabular}
%     }
%     \caption{Results on , the \all~setting provides the best performance with BertScore (bolded) for each model --- differences in performance are significant for $p < 0.0001$. } 
%     \label{tab:dev-components}
%     \vspace{-10pt}
% \end{table*}

\paragraph{Automatic parsing is good enough}

Coptic-NLP provides high quality parses (\cref{sec:coptic-nlp}) with a labeled attachment score of 89.7 \cite{zeldes-etal-2025-ud}. These high-quality automatic parses are useful and effective. But are they as useful as gold standard parses?

In general, we observe performance improvements with both gold and automatic parses, with no consistent difference between their benefits. Although we assume that parsing accuracy should make a difference, for our data and samples we find no systematic impact on downstream translation quality, suggesting that automatic parsing errors from a $\sim$90\% accurate parse may not cascade into drastic translation errors. \Cref{tab:dev-gold-highlights} shows the performance of the syntax settings of the open models (for \dev) with automatic and gold parses.
\\
\paragraph{Biblical text fares better}

\begin{table}[!t]
    \centering
    \begin{tabular}{lrrrr}
    & \multicolumn{2}{c}{\textbf{Gemma-12b}} &  \multicolumn{2}{c}{\textbf{Gemma-27b}}\\
    \midrule
     &  \textbf{Bible}&  \textbf{Other} &  \textbf{Bible}&  \textbf{Other} \\
    \midrule \textbf{Setting}     & \multicolumn{4}{c}{\textbf{BertScore}}\\
    \midrule
Baseline & {0.8323} & 0.8297 &  {0.8378} &0.8361 \\
 \conll & {0.8458} & 0.8436 &  {0.8519} &0.8482 \\
 \con & {0.8537} & 0.8448 &  {0.8544} &0.8453 \\
 \dep & {0.8422} & 0.8373 &  {0.8423} &0.8390 \\
 \lexicon & {0.8585} & 0.8582 &  {0.8635} &0.8577 \\
 \dep+\con & {0.8496} & 0.8467 &  {0.8538} &0.8455 \\
 \all & {0.8738} & 0.8685 &  {0.8792} &0.8665 \\
 \bottomrule
    \end{tabular}
    \caption{Results of Gemma model for \dev~reported for based on whether they are from the Bible (182 of 380). The BertScore in each setting (including baseline) is higher for the Biblical text. Results with BertScore and BLEU are reported in appendix (\cref{tab:dev-bible}).}
    \label{tab:dev-bible-highlights}
\end{table}

Models may perform better in translating Biblical text than non-Biblical text~\cite{liu-etal-2021-usefulness}. We see that to be the case for our \dev~set~(\cref{tab:dev-bible-highlights}). All settings do better for the Bible subset, to varying degrees.

It is also not hard to find qualitative examples in which baseline GPT-4.1 results strongly suggest recognition of Bible content. For example, for the reference target in \cref{ex:ephthaha}~(Mark 7:34), the baseline GPT prediction in \cref{ex:ephthaha-gpt} is very good and gets `be opened' right, likely because the Greek letters rendering the foreign term `Ephphatha' give away the related Bible verse, which the model has memorized.

\ex. Looking up to heaven, he sighed, and said to him, `Ephphatha!' that is, `Be opened!'~\text{(Reference)}\label{ex:ephthaha}

\ex. And when he had come near, he touched him and said to him,``Be opened'' And immediately it was opened. (GPT-4.1)\label{ex:ephthaha-gpt}

\section{Conclusion}

We propose augmenting in-context translation for low-resource languages with syntactic information. Such augmentation could support better translation compared to just dictionary-based augmentation.

We build components for dictionary and multiple operationalizations of syntactic augmentation based on Universal Dependencies parses, for in-context translation of the low-resource language of Coptic. We validate these components with multiple open-weight (mainly Gemma) and reference closed-source (mainly GPT-4.1) models, and report results for both in-domain and out-of-domain evaluation sets.

Inclusion of syntactic information in addition to dictionary information provides higher quality translations than just supplying dictionary information, leading to statistically significant improvements in BERTScore F1 and observable improvements in translation quality. Gains in out-of-domain data for can also be seen in scores compared to previously reported results on Coptic ostraca.

We show that even limited linguistic resources can meaningfully improve in-context translation quality for low-resource languages, with improvements seen in both open and closed models. 

The type of added syntactic information that is useful can vary by model: while larger models such as GPT-4.1 can make more effective use of UD parses in a raw, CoNLL-U format, smaller models benefit more from verbalizations. All models exhibit improvements from a dedicated construction module explaining constructions specific to the Coptic language and pointing out tricky configurations, as well as indicating and transliterating proper names. For GPT-4.1 this benefit emerges primarily when combined with raw parses.

Additionally, we also find minimal differences between gold and automatic UD parses, showing that high-quality automatic parses are sufficient for in-context translation, at least given parser performance of $\sim$90\%. Our error analysis highlights how different information can provide support for the models to identify both structural and relational information to improve translations. 

We are hopeful that these results will lead to more work on incorporating abstract syntactic or even semantic information in general (for example WordNet or similar resources, \citealt{slaughter-etal-2019-making}, or semantic analysis graphs such as UMR, \citealt{vangysel2021umr}), and outputs from automatic UD parsing in particular, into ICL approaches to low-resource machine translation.

\section*{Limitations}
\paragraph{Small dataset} Our experiments are on a dataset of less than 800 translation samples. This limits the generalizability of our findings.

\paragraph{Single operationalization of lexicon} We consider dictionary augmentation with a single source, and one operationalization, based on the resource we use (although aligned with TEI standard). However, dictionaries can differ and consequently augmentation may not be adaptable.

\paragraph{Single formalism} While Universal Dependencies is both the most popular and diverse (across languages) collection of treebanks, it is not the only syntactic or grammatical formalism available. However, the verbalizations are more generic, and could help mitigate this limitation. 

\paragraph{Non exhaustive settings combination} We selected a subset of experimental settings, and some combinations of linguistic components were not included, such as \dep+\conll.

\paragraph{Prompt length} Our settings incorporate multiple linguistic components, producing long prompts in some cases. The length of these prompts can provide additional challenges for LLMs, as they are sensitive to context length, possibly affecting translation quality for more complex experimental settings. 

\paragraph{Unattested metrics}Coptic, being a low-resource language, has not had extensive investigation of MT metrics and their relation to human judgment. Although our target language is English, there is no attestation of metrics for Coptic-English translation. Hence the reported results may not represent human judgments of translation quality. 

\paragraph{Lack of human evaluation} We do not perform any human evaluation, but rather only use automatic metrics based on human-translated references.

\section*{Acknowledgments}
The work in this paper was supported by the Georgetown University Massive Data Institute (MDI) through funding for an MDI Scholar. Compute and inference resources provided through the Department of Linguistics and the Massive Data Institute at Georgetown University were used for this work.

% Bibliography entries for the entire Anthology, followed by custom entries
%\bibliography{anthology,custom}
% Custom bibliography entries only
\bibliography{custom}

@inproceedings{miyagawa-2025-rag,
    title = "{RAG}-Enhanced Neural Machine Translation of {A}ncient {E}gyptian Text: A Case Study of {THOTH} {AI}",
    author = "Miyagawa, So",
    editor = {H{\"a}m{\"a}l{\"a}inen, Mika  and
      {\"O}hman, Emily  and
      Bizzoni, Yuri  and
      Miyagawa, So  and
      Alnajjar, Khalid},
    booktitle = "Proceedings of the 5th International Conference on Natural Language Processing for Digital Humanities",
    month = may,
    year = "2025",
    address = "Albuquerque, USA",
    publisher = "Association for Computational Linguistics",
    url = "https://aclanthology.org/2025.nlp4dh-1.4/",
    doi = "10.18653/v1/2025.nlp4dh-1.4",
    pages = "33--40",
    ISBN = "979-8-89176-234-3"
}

@inproceedings{saeed-etal-2024-nile,
    title = "From {N}ile Sands to Digital Hands: Machine Translation of {C}optic Texts",
    author = "Saeed, Muhammed  and
      Mohamed, Asim  and
      Mohamed, Mukhtar  and
      Shehata, Shady  and
      Abdul-Mageed, Muhammad",
    editor = "Habash, Nizar  and
      Bouamor, Houda  and
      Eskander, Ramy  and
      Tomeh, Nadi  and
      Abu Farha, Ibrahim  and
      Abdelali, Ahmed  and
      Touileb, Samia  and
      Hamed, Injy  and
      Onaizan, Yaser  and
      Alhafni, Bashar  and
      Antoun, Wissam  and
      Khalifa, Salam  and
      Haddad, Hatem  and
      Zitouni, Imed  and
      AlKhamissi, Badr  and
      Almatham, Rawan  and
      Mrini, Khalil",
    booktitle = "Proceedings of the Second Arabic Natural Language Processing Conference",
    month = aug,
    year = "2024",
    address = "Bangkok, Thailand",
    publisher = "Association for Computational Linguistics",
    url = "https://aclanthology.org/2024.arabicnlp-1.25/",
    doi = "10.18653/v1/2024.arabicnlp-1.25",
    pages = "298--308"
}

@inproceedings{wannaz-miyagawa-2024-assessing,
    title = "Assessing Large Language Models in Translating {C}optic and {A}ncient {G}reek Ostraca",
    author = "Wannaz, Audric-Charles  and
      Miyagawa, So",
    editor = {H{\"a}m{\"a}l{\"a}inen, Mika  and
      {\"O}hman, Emily  and
      Miyagawa, So  and
      Alnajjar, Khalid  and
      Bizzoni, Yuri},
    booktitle = "Proceedings of the 4th International Conference on Natural Language Processing for Digital Humanities",
    month = nov,
    year = "2024",
    address = "Miami, USA",
    publisher = "Association for Computational Linguistics",
    url = "https://aclanthology.org/2024.nlp4dh-1.44/",
    doi = "10.18653/v1/2024.nlp4dh-1.44",
    pages = "463--471"
}

@misc{chaoui2025neuralmachinetranslationcopticfrench,
      title={Neural Machine Translation for {C}optic-{F}rench: Strategies for Low-Resource Ancient Languages}, 
      author={Nasma Chaoui and Richard Khoury},
      year={2025},
      eprint={2508.10683},
      archivePrefix={arXiv},
      primaryClass={cs.CL},
      url={https://arxiv.org/abs/2508.10683}, 
}

@article{de-marneffe-etal-2021-universal,
    title = "{U}niversal {D}ependencies",
    author = "de Marneffe, Marie-Catherine  and
      Manning, Christopher D.  and
      Nivre, Joakim  and
      Zeman, Daniel",
    journal = "Computational Linguistics",
    volume = "47",
    number = "2",
    month = jun,
    year = "2021",
    address = "Cambridge, MA",
    publisher = "MIT Press",
    url = "https://aclanthology.org/2021.cl-2.11/",
    doi = "10.1162/coli_a_00402",
    pages = "255--308"
}

@Article{SchroederZeldes2015,
  author    = {Caroline T. Schroeder and Amir Zeldes},
  title     = {Raiders of the Lost Corpus},
  journal   = {Digital Humanities Quarterly},
  year      = {2016},
  volume    = {10},
  number    = {2},
  url       = {http://digitalhumanities.org/dhq/vol/10/2/000247/000247.html},
}

@inproceedings{zeldes-abrams-2018-coptic,
    title = "The {C}optic {U}niversal {D}ependency Treebank",
    author = "Zeldes, Amir  and
      Abrams, Mitchell",
    editor = "de Marneffe, Marie-Catherine  and
      Lynn, Teresa  and
      Schuster, Sebastian",
    booktitle = "Proceedings of the Second Workshop on Universal Dependencies ({UDW} 2018)",
    month = nov,
    year = "2018",
    address = "Brussels, Belgium",
    publisher = "Association for Computational Linguistics",
    url = "https://aclanthology.org/W18-6022/",
    doi = "10.18653/v1/W18-6022",
    pages = "192--201"
}

@inproceedings{feder-etal-2018-linked,
    title = "A Linked {C}optic Dictionary Online",
    author = "Feder, Frank  and
      Kupreyev, Maxim  and
      Manning, Emma  and
      Schroeder, Caroline T.  and
      Zeldes, Amir",
    editor = "Alex, Beatrice  and
      Degaetano-Ortlieb, Stefania  and
      Feldman, Anna  and
      Kazantseva, Anna  and
      Reiter, Nils  and
      Szpakowicz, Stan",
    booktitle = "Proceedings of the Second Joint {SIGHUM} Workshop on Computational Linguistics for Cultural Heritage, Social Sciences, Humanities and Literature",
    month = aug,
    year = "2018",
    address = "Santa Fe, New Mexico",
    publisher = "Association for Computational Linguistics",
    url = "https://aclanthology.org/W18-4502/",
    pages = "12--21"
}

@inproceedings{wmt25-instructions,
    title = "Findings of the {WMT}25 Multilingual Instruction Shared Task: Persistent Hurdles in Reasoning, Generation, and Evaluation",
    author = "Kocmi, Tom  and
      Agrawal, Sweta  and
      Artemova, Ekaterina  and
      Avramidis, Eleftherios  and
      Briakou, Eleftheria  and
      Chen, Pinzhen  and
      Fadaee, Marzieh  and
      Freitag, Markus  and
      Grundkiewicz, Roman  and
      Hou, Yupeng  and
      Koehn, Philipp  and
      Kreutzer, Julia  and
      Mansour, Saab  and
      Perrella, Stefano  and
      Proietti, Lorenzo  and
      Riley, Parker  and
      S{\'a}nchez, Eduardo  and
      Schmidtova, Patricia  and
      Shmatova, Mariya  and
      Zouhar, Vil{\'e}m",
    editor = "Haddow, Barry  and
      Kocmi, Tom  and
      Koehn, Philipp  and
      Monz, Christof",
    booktitle = "Proceedings of the Tenth Conference on Machine Translation",
    month = nov,
    year = "2025",
    address = "Suzhou, China",
    publisher = "Association for Computational Linguistics",
    url = "https://aclanthology.org/2025.wmt-1.23/",
    doi = "10.18653/v1/2025.wmt-1.23",
    pages = "414--435",
    ISBN = "979-8-89176-341-8"
}

@inproceedings{kocmi-etal-2025-findings,
    title = "Findings of the {WMT}25 General Machine Translation Shared Task: Time to Stop Evaluating on Easy Test Sets",
    author = "Kocmi, Tom  and
      Artemova, Ekaterina  and
      Avramidis, Eleftherios  and
      Bawden, Rachel  and
      Bojar, Ond{\v{r}}ej  and
      Dranch, Konstantin  and
      Dvorkovich, Anton  and
      Dukanov, Sergey  and
      Fishel, Mark  and
      Freitag, Markus  and
      Gowda, Thamme  and
      Grundkiewicz, Roman  and
      Haddow, Barry  and
      Karpinska, Marzena  and
      Koehn, Philipp  and
      Lakougna, Howard  and
      Lundin, Jessica  and
      Monz, Christof  and
      Murray, Kenton  and
      Nagata, Masaaki  and
      Perrella, Stefano  and
      Proietti, Lorenzo  and
      Popel, Martin  and
      Popovi{\'c}, Maja  and
      Riley, Parker  and
      Shmatova, Mariya  and
      Steingr{\'i}msson, Steinth{\'o}r  and
      Yankovskaya, Lisa  and
      Zouhar, Vil{\'e}m",
    editor = "Haddow, Barry  and
      Kocmi, Tom  and
      Koehn, Philipp  and
      Monz, Christof",
    booktitle = "Proceedings of the Tenth Conference on Machine Translation",
    month = nov,
    year = "2025",
    address = "Suzhou, China",
    publisher = "Association for Computational Linguistics",
    url = "https://aclanthology.org/2025.wmt-1.22/",
    doi = "10.18653/v1/2025.wmt-1.22",
    pages = "355--413",
    ISBN = "979-8-89176-341-8"
}

@inproceedings{nagy-etal-2023-treeswap,
    title = "{T}ree{S}wap: Data Augmentation for Machine Translation via Dependency Subtree Swapping",
    author = "Nagy, Attila  and
      Lakatos, Dorina  and
      Barta, Botond  and
      {\'A}cs, Judit",
    editor = "Mitkov, Ruslan  and
      Angelova, Galia",
    booktitle = "Proceedings of the 14th International Conference on Recent Advances in Natural Language Processing",
    month = sep,
    year = "2023",
    address = "Varna, Bulgaria",
    publisher = "INCOMA Ltd., Shoumen, Bulgaria",
    url = "https://aclanthology.org/2023.ranlp-1.82/",
    pages = "759--768"
}

@inproceedings{pei-etal-2025-understanding,
    title = "Understanding In-Context Machine Translation for Low-Resource Languages: A Case Study on {M}anchu",
    author = "Pei, Renhao  and
      Liu, Yihong  and
      Lin, Peiqin  and
      Yvon, Fran{\c{c}}ois  and
      Schuetze, Hinrich",
    editor = "Che, Wanxiang  and
      Nabende, Joyce  and
      Shutova, Ekaterina  and
      Pilehvar, Mohammad Taher",
    booktitle = "Proceedings of the 63rd Annual Meeting of the Association for Computational Linguistics (Volume 1: Long Papers)",
    month = jul,
    year = "2025",
    address = "Vienna, Austria",
    publisher = "Association for Computational Linguistics",
    url = "https://aclanthology.org/2025.acl-long.429/",
    doi = "10.18653/v1/2025.acl-long.429",
    pages = "8767--8788",
    ISBN = "979-8-89176-251-0"
}

@inproceedings{peng-zeldes-2018-roads,
    IDS = {depedit},
    title = "All Roads Lead to {UD}: Converting {S}tanford and {P}enn Parses to {E}nglish {U}niversal {D}ependencies with Multilayer Annotations",
    author = "Peng, Siyao  and
      Zeldes, Amir",
    editor = "Savary, Agata  and
      Ramisch, Carlos  and
      Hwang, Jena D.  and
      Schneider, Nathan  and
      Andresen, Melanie  and
      Pradhan, Sameer  and
      Petruck, Miriam R. L.",
    booktitle = "Proceedings of the Joint Workshop on Linguistic Annotation, Multiword Expressions and Constructions ({LAW}-{MWE}-{C}x{G}-2018)",
    month = aug,
    year = "2018",
    address = "Santa Fe, New Mexico, USA",
    publisher = "Association for Computational Linguistics",
    url = "https://aclanthology.org/W18-4918/",
    pages = "167--177"
}

@inproceedings{bertscore,
    title={BERTScore: Evaluating Text Generation with BERT},
    author={Tianyi Zhang and Varsha Kishore and Felix Wu and Kilian Q. Weinberger and Yoav Artzi},
    booktitle={International Conference on Learning Representations},
    year={2020},
    url={https://openreview.net/forum?id=SkeHuCVFDr}
}

@inproceedings{lavie-etal-2025-findings,
    title = "Findings of the {WMT}25 Shared Task on Automated Translation Evaluation Systems: Linguistic Diversity is Challenging and References Still Help",
    author = "Lavie, Alon  and
      Hanneman, Greg  and
      Agrawal, Sweta  and
      Kanojia, Diptesh  and
      Lo, Chi-Kiu  and
      Zouhar, Vil{\'e}m  and
      Blain, Frederic  and
      Zerva, Chrysoula  and
      Avramidis, Eleftherios  and
      Deoghare, Sourabh  and
      Sindhujan, Archchana  and
      Wang, Jiayi  and
      Adelani, David Ifeoluwa  and
      Thompson, Brian  and
      Kocmi, Tom  and
      Freitag, Markus  and
      Deutsch, Daniel",
    editor = "Haddow, Barry  and
      Kocmi, Tom  and
      Koehn, Philipp  and
      Monz, Christof",
    booktitle = "Proceedings of the Tenth Conference on Machine Translation",
    month = nov,
    year = "2025",
    address = "Suzhou, China",
    publisher = "Association for Computational Linguistics",
    url = "https://aclanthology.org/2025.wmt-1.24/",
    doi = "10.18653/v1/2025.wmt-1.24",
    pages = "436--483",
    ISBN = "979-8-89176-341-8"
}

@inproceedings{zeldes-schroeder-2016-nlp,
    title = "An {NLP} Pipeline for {C}optic",
    author = "Zeldes, Amir  and
      Schroeder, Caroline T.",
    editor = "Reiter, Nils  and
      Alex, Beatrice  and
      Zervanou, Kalliopi A.",
    booktitle = "Proceedings of the 10th {SIGHUM} Workshop on Language Technology for Cultural Heritage, Social Sciences, and Humanities",
    month = aug,
    year = "2016",
    address = "Berlin, Germany",
    publisher = "Association for Computational Linguistics",
    url = "https://aclanthology.org/W16-2119/",
    doi = "10.18653/v1/W16-2119",
    pages = "146--155"
}

@inproceedings{lu-etal-2024-chain,
    title = "Chain-of-Dictionary Prompting Elicits Translation in Large Language Models",
    author = "Lu, Hongyuan  and
      Yang, Haoran  and
      Huang, Haoyang  and
      Zhang, Dongdong  and
      Lam, Wai  and
      Wei, Furu",
    editor = "Al-Onaizan, Yaser  and
      Bansal, Mohit  and
      Chen, Yun-Nung",
    booktitle = "Proceedings of the 2024 Conference on Empirical Methods in Natural Language Processing",
    month = nov,
    year = "2024",
    address = "Miami, Florida, USA",
    publisher = "Association for Computational Linguistics",
    url = "https://aclanthology.org/2024.emnlp-main.55/",
    doi = "10.18653/v1/2024.emnlp-main.55",
    pages = "958--976"
}

@misc{ghazvininejad-et-al-2023-dictionary,
      title={Dictionary-based Phrase-level Prompting of Large Language Models for Machine Translation}, 
      author={Marjan Ghazvininejad and Hila Gonen and Luke Zettlemoyer},
      year={2023},
      eprint={2302.07856},
      archivePrefix={arXiv},
      primaryClass={cs.CL},
      url={https://arxiv.org/abs/2302.07856}, 
}

@inproceedings{elsner-needle-2023-translating,
    title = "Translating a low-resource language using {GPT}-3 and a human-readable dictionary",
    author = "Elsner, Micha  and
      Needle, Jordan",
    editor = {Nicolai, Garrett  and
      Chodroff, Eleanor  and
      Mailhot, Frederic  and
      {\c{C}}{\"o}ltekin, {\c{C}}a{\u{g}}r{\i}},
    booktitle = "Proceedings of the 20th SIGMORPHON workshop on Computational Research in Phonetics, Phonology, and Morphology",
    month = jul,
    year = "2023",
    address = "Toronto, Canada",
    publisher = "Association for Computational Linguistics",
    url = "https://aclanthology.org/2023.sigmorphon-1.2/",
    doi = "10.18653/v1/2023.sigmorphon-1.2",
    pages = "1--13"
}

@inproceedings{court-elsner-2024-shortcomings,
    title = "Shortcomings of {LLM}s for Low-Resource Translation: Retrieval and Understanding Are Both the Problem",
    author = "Court, Sara  and
      Elsner, Micha",
    editor = "Haddow, Barry  and
      Kocmi, Tom  and
      Koehn, Philipp  and
      Monz, Christof",
    booktitle = "Proceedings of the Ninth Conference on Machine Translation",
    month = nov,
    year = "2024",
    address = "Miami, Florida, USA",
    publisher = "Association for Computational Linguistics",
    url = "https://aclanthology.org/2024.wmt-1.125/",
    doi = "10.18653/v1/2024.wmt-1.125",
    pages = "1332--1354"
}

@article{liu-etal-2024-lost,
	title        = {Lost in the Middle: How Language Models Use Long Contexts},
	author       = {Liu, Nelson F.  and       Lin, Kevin  and       Hewitt, John  and       Paranjape, Ashwin  and       Bevilacqua, Michele  and       Petroni, Fabio  and       Liang, Percy},
	year         = 2024,
	journal      = {Transactions of the Association for Computational Linguistics},
	publisher    = {MIT Press},
	address      = {Cambridge, MA},
	volume       = 12,
	pages        = {157--173},
	doi          = {10.1162/tacl_a_00638},
	url          = {https://aclanthology.org/2024.tacl-1.9/}
}

@inproceedings{tanzer-2024-mtob,
	title        = {A Benchmark for Learning to Translate a New Language from One Grammar Book},
	author       = {Garrett Tanzer and Mirac Suzgun and Eline Visser and Dan Jurafsky and Luke Melas-Kyriazi},
	year         = 2024,
	booktitle    = {The Twelfth International Conference on Learning Representations},
	url          = {https://openreview.net/forum?id=tbVWug9f2h}
}

@inproceedings{zhang-etal-2025-read,
    title = "Read it in Two Steps: Translating Extremely Low-Resource Languages with Code-Augmented Grammar Books",
    author = "Zhang, Chen  and
      Lin, Jiuheng  and
      Liu, Xiao  and
      Zhang, Zekai  and
      Feng, Yansong",
    editor = "Che, Wanxiang  and
      Nabende, Joyce  and
      Shutova, Ekaterina  and
      Pilehvar, Mohammad Taher",
    booktitle = "Proceedings of the 63rd Annual Meeting of the Association for Computational Linguistics (Volume 1: Long Papers)",
    month = jul,
    year = "2025",
    address = "Vienna, Austria",
    publisher = "Association for Computational Linguistics",
    url = "https://aclanthology.org/2025.acl-long.202/",
    doi = "10.18653/v1/2025.acl-long.202",
    pages = "3977--3997",
    ISBN = "979-8-89176-251-0"
}

@misc{gpt4technicalreport,
      title={GPT-4 Technical Report}, 
      author={OpenAI and Josh Achiam and Steven Adler and Sandhini Agarwal and Lama Ahmad and Ilge Akkaya and Florencia Leoni Aleman and Diogo Almeida and Janko Altenschmidt and Sam Altman and Shyamal Anadkat and Red Avila and Igor Babuschkin and Suchir Balaji and Valerie Balcom and Paul Baltescu and Haiming Bao and Mohammad Bavarian and Jeff Belgum and Irwan Bello and Jake Berdine and Gabriel Bernadett-Shapiro and Christopher Berner and Lenny Bogdonoff and Oleg Boiko and Madelaine Boyd and Anna-Luisa Brakman and Greg Brockman and Tim Brooks and Miles Brundage and Kevin Button and Trevor Cai and Rosie Campbell and Andrew Cann and Brittany Carey and Chelsea Carlson and Rory Carmichael and Brooke Chan and Che Chang and Fotis Chantzis and Derek Chen and Sully Chen and Ruby Chen and Jason Chen and Mark Chen and Ben Chess and Chester Cho and Casey Chu and Hyung Won Chung and Dave Cummings and Jeremiah Currier and Yunxing Dai and Cory Decareaux and Thomas Degry and Noah Deutsch and Damien Deville and Arka Dhar and David Dohan and Steve Dowling and Sheila Dunning and Adrien Ecoffet and Atty Eleti and Tyna Eloundou and David Farhi and Liam Fedus and Niko Felix and Simón Posada Fishman and Juston Forte and Isabella Fulford and Leo Gao and Elie Georges and Christian Gibson and Vik Goel and Tarun Gogineni and Gabriel Goh and Rapha Gontijo-Lopes and Jonathan Gordon and Morgan Grafstein and Scott Gray and Ryan Greene and Joshua Gross and Shixiang Shane Gu and Yufei Guo and Chris Hallacy and Jesse Han and Jeff Harris and Yuchen He and Mike Heaton and Johannes Heidecke and Chris Hesse and Alan Hickey and Wade Hickey and Peter Hoeschele and Brandon Houghton and Kenny Hsu and Shengli Hu and Xin Hu and Joost Huizinga and Shantanu Jain and Shawn Jain and Joanne Jang and Angela Jiang and Roger Jiang and Haozhun Jin and Denny Jin and Shino Jomoto and Billie Jonn and Heewoo Jun and Tomer Kaftan and Łukasz Kaiser and Ali Kamali and Ingmar Kanitscheider and Nitish Shirish Keskar and Tabarak Khan and Logan Kilpatrick and Jong Wook Kim and Christina Kim and Yongjik Kim and Jan Hendrik Kirchner and Jamie Kiros and Matt Knight and Daniel Kokotajlo and Łukasz Kondraciuk and Andrew Kondrich and Aris Konstantinidis and Kyle Kosic and Gretchen Krueger and Vishal Kuo and Michael Lampe and Ikai Lan and Teddy Lee and Jan Leike and Jade Leung and Daniel Levy and Chak Ming Li and Rachel Lim and Molly Lin and Stephanie Lin and Mateusz Litwin and Theresa Lopez and Ryan Lowe and Patricia Lue and Anna Makanju and Kim Malfacini and Sam Manning and Todor Markov and Yaniv Markovski and Bianca Martin and Katie Mayer and Andrew Mayne and Bob McGrew and Scott Mayer McKinney and Christine McLeavey and Paul McMillan and Jake McNeil and David Medina and Aalok Mehta and Jacob Menick and Luke Metz and Andrey Mishchenko and Pamela Mishkin and Vinnie Monaco and Evan Morikawa and Daniel Mossing and Tong Mu and Mira Murati and Oleg Murk and David Mély and Ashvin Nair and Reiichiro Nakano and Rajeev Nayak and Arvind Neelakantan and Richard Ngo and Hyeonwoo Noh and Long Ouyang and Cullen O'Keefe and Jakub Pachocki and Alex Paino and Joe Palermo and Ashley Pantuliano and Giambattista Parascandolo and Joel Parish and Emy Parparita and Alex Passos and Mikhail Pavlov and Andrew Peng and Adam Perelman and Filipe de Avila Belbute Peres and Michael Petrov and Henrique Ponde de Oliveira Pinto and Michael and Pokorny and Michelle Pokrass and Vitchyr H. Pong and Tolly Powell and Alethea Power and Boris Power and Elizabeth Proehl and Raul Puri and Alec Radford and Jack Rae and Aditya Ramesh and Cameron Raymond and Francis Real and Kendra Rimbach and Carl Ross and Bob Rotsted and Henri Roussez and Nick Ryder and Mario Saltarelli and Ted Sanders and Shibani Santurkar and Girish Sastry and Heather Schmidt and David Schnurr and John Schulman and Daniel Selsam and Kyla Sheppard and Toki Sherbakov and Jessica Shieh and Sarah Shoker and Pranav Shyam and Szymon Sidor and Eric Sigler and Maddie Simens and Jordan Sitkin and Katarina Slama and Ian Sohl and Benjamin Sokolowsky and Yang Song and Natalie Staudacher and Felipe Petroski Such and Natalie Summers and Ilya Sutskever and Jie Tang and Nikolas Tezak and Madeleine B. Thompson and Phil Tillet and Amin Tootoonchian and Elizabeth Tseng and Preston Tuggle and Nick Turley and Jerry Tworek and Juan Felipe Cerón Uribe and Andrea Vallone and Arun Vijayvergiya and Chelsea Voss and Carroll Wainwright and Justin Jay Wang and Alvin Wang and Ben Wang and Jonathan Ward and Jason Wei and CJ Weinmann and Akila Welihinda and Peter Welinder and Jiayi Weng and Lilian Weng and Matt Wiethoff and Dave Willner and Clemens Winter and Samuel Wolrich and Hannah Wong and Lauren Workman and Sherwin Wu and Jeff Wu and Michael Wu and Kai Xiao and Tao Xu and Sarah Yoo and Kevin Yu and Qiming Yuan and Wojciech Zaremba and Rowan Zellers and Chong Zhang and Marvin Zhang and Shengjia Zhao and Tianhao Zheng and Juntang Zhuang and William Zhuk and Barret Zoph},
      year={2024},
      eprint={2303.08774},
      archivePrefix={arXiv},
      primaryClass={cs.CL},
      url={https://arxiv.org/abs/2303.08774}, 
}

@misc{gemma3technicalreport,       title={Gemma 3 Technical Report},        author={Gemma Team and Aishwarya Kamath and Johan Ferret and Shreya Pathak and Nino Vieillard and Ramona Merhej and Sarah Perrin and Tatiana Matejovicova and Alexandre Ramé and Morgane Rivière and Louis Rouillard and Thomas Mesnard and Geoffrey Cideron and Jean-bastien Grill and Sabela Ramos and Edouard Yvinec and Michelle Casbon and Etienne Pot and Ivo Penchev and Gaël Liu and Francesco Visin and Kathleen Kenealy and Lucas Beyer and Xiaohai Zhai and Anton Tsitsulin and Robert Busa-Fekete and Alex Feng and Noveen Sachdeva and Benjamin Coleman and Yi Gao and Basil Mustafa and Iain Barr and Emilio Parisotto and David Tian and Matan Eyal and Colin Cherry and Jan-Thorsten Peter and Danila Sinopalnikov and Surya Bhupatiraju and Rishabh Agarwal and Mehran Kazemi and Dan Malkin and Ravin Kumar and David Vilar and Idan Brusilovsky and Jiaming Luo and Andreas Steiner and Abe Friesen and Abhanshu Sharma and Abheesht Sharma and Adi Mayrav Gilady and Adrian Goedeckemeyer and Alaa Saade and Alex Feng and Alexander Kolesnikov and Alexei Bendebury and Alvin Abdagic and Amit Vadi and András György and André Susano Pinto and Anil Das and Ankur Bapna and Antoine Miech and Antoine Yang and Antonia Paterson and Ashish Shenoy and Ayan Chakrabarti and Bilal Piot and Bo Wu and Bobak Shahriari and Bryce Petrini and Charlie Chen and Charline Le Lan and Christopher A. Choquette-Choo and CJ Carey and Cormac Brick and Daniel Deutsch and Danielle Eisenbud and Dee Cattle and Derek Cheng and Dimitris Paparas and Divyashree Shivakumar Sreepathihalli and Doug Reid and Dustin Tran and Dustin Zelle and Eric Noland and Erwin Huizenga and Eugene Kharitonov and Frederick Liu and Gagik Amirkhanyan and Glenn Cameron and Hadi Hashemi and Hanna Klimczak-Plucińska and Harman Singh and Harsh Mehta and Harshal Tushar Lehri and Hussein Hazimeh and Ian Ballantyne and Idan Szpektor and Ivan Nardini and Jean Pouget-Abadie and Jetha Chan and Joe Stanton and John Wieting and Jonathan Lai and Jordi Orbay and Joseph Fernandez and Josh Newlan and Ju-yeong Ji and Jyotinder Singh and Kat Black and Kathy Yu and Kevin Hui and Kiran Vodrahalli and Klaus Greff and Linhai Qiu and Marcella Valentine and Marina Coelho and Marvin Ritter and Matt Hoffman and Matthew Watson and Mayank Chaturvedi and Michael Moynihan and Min Ma and Nabila Babar and Natasha Noy and Nathan Byrd and Nick Roy and Nikola Momchev and Nilay Chauhan and Noveen Sachdeva and Oskar Bunyan and Pankil Botarda and Paul Caron and Paul Kishan Rubenstein and Phil Culliton and Philipp Schmid and Pier Giuseppe Sessa and Pingmei Xu and Piotr Stanczyk and Pouya Tafti and Rakesh Shivanna and Renjie Wu and Renke Pan and Reza Rokni and Rob Willoughby and Rohith Vallu and Ryan Mullins and Sammy Jerome and Sara Smoot and Sertan Girgin and Shariq Iqbal and Shashir Reddy and Shruti Sheth and Siim Põder and Sijal Bhatnagar and Sindhu Raghuram Panyam and Sivan Eiger and Susan Zhang and Tianqi Liu and Trevor Yacovone and Tyler Liechty and Uday Kalra and Utku Evci and Vedant Misra and Vincent Roseberry and Vlad Feinberg and Vlad Kolesnikov and Woohyun Han and Woosuk Kwon and Xi Chen and Yinlam Chow and Yuvein Zhu and Zichuan Wei and Zoltan Egyed and Victor Cotruta and Minh Giang and Phoebe Kirk and Anand Rao and Kat Black and Nabila Babar and Jessica Lo and Erica Moreira and Luiz Gustavo Martins and Omar Sanseviero and Lucas Gonzalez and Zach Gleicher and Tris Warkentin and Vahab Mirrokni and Evan Senter and Eli Collins and Joelle Barral and Zoubin Ghahramani and Raia Hadsell and Yossi Matias and D. Sculley and Slav Petrov and Noah Fiedel and Noam Shazeer and Oriol Vinyals and Jeff Dean and Demis Hassabis and Koray Kavukcuoglu and Clement Farabet and Elena Buchatskaya and Jean-Baptiste Alayrac and Rohan Anil and Dmitry and Lepikhin and Sebastian Borgeaud and Olivier Bachem and Armand Joulin and Alek Andreev and Cassidy Hardin and Robert Dadashi and Léonard Hussenot},       year={2025},       eprint={2503.19786},       archivePrefix={arXiv},       primaryClass={cs.CL},       url={https://arxiv.org/abs/2503.19786},  }

@inproceedings{papineni-etal-2002-bleu,
    title = "{B}leu: a Method for Automatic Evaluation of Machine Translation",
    author = "Papineni, Kishore  and
      Roukos, Salim  and
      Ward, Todd  and
      Zhu, Wei-Jing",
    editor = "Isabelle, Pierre  and
      Charniak, Eugene  and
      Lin, Dekang",
    booktitle = "Proceedings of the 40th Annual Meeting of the Association for Computational Linguistics",
    month = jul,
    year = "2002",
    address = "Philadelphia, Pennsylvania, USA",
    publisher = "Association for Computational Linguistics",
    url = "https://aclanthology.org/P02-1040/",
    doi = "10.3115/1073083.1073135",
    pages = "311--318"
}

@inproceedings{popovic-2017-chrf,
    title = "chr{F}++: words helping character n-grams",
    author = "Popovi{\'c}, Maja",
    editor = "Bojar, Ond{\v{r}}ej  and
      Buck, Christian  and
      Chatterjee, Rajen  and
      Federmann, Christian  and
      Graham, Yvette  and
      Haddow, Barry  and
      Huck, Matthias  and
      Yepes, Antonio Jimeno  and
      Koehn, Philipp  and
      Kreutzer, Julia",
    booktitle = "Proceedings of the Second Conference on Machine Translation",
    month = sep,
    year = "2017",
    address = "Copenhagen, Denmark",
    publisher = "Association for Computational Linguistics",
    url = "https://aclanthology.org/W17-4770/",
    doi = "10.18653/v1/W17-4770",
    pages = "612--618"
}

@inproceedings{banerjee-lavie-2005-meteor,
    title = "{METEOR}: An Automatic Metric for {MT} Evaluation with Improved Correlation with Human Judgments",
    author = "Banerjee, Satanjeev  and
      Lavie, Alon",
    editor = "Goldstein, Jade  and
      Lavie, Alon  and
      Lin, Chin-Yew  and
      Voss, Clare",
    booktitle = "Proceedings of the {ACL} Workshop on Intrinsic and Extrinsic Evaluation Measures for Machine Translation and/or Summarization",
    month = jun,
    year = "2005",
    address = "Ann Arbor, Michigan",
    publisher = "Association for Computational Linguistics",
    url = "https://aclanthology.org/W05-0909/",
    pages = "65--72"
}

@inproceedings{post-2018-call,
    title = "A Call for Clarity in Reporting {BLEU} Scores",
    author = "Post, Matt",
    editor = "Bojar, Ond{\v{r}}ej  and
      Chatterjee, Rajen  and
      Federmann, Christian  and
      Fishel, Mark  and
      Graham, Yvette  and
      Haddow, Barry  and
      Huck, Matthias  and
      Yepes, Antonio Jimeno  and
      Koehn, Philipp  and
      Monz, Christof  and
      Negri, Matteo  and
      N{\'e}v{\'e}ol, Aur{\'e}lie  and
      Neves, Mariana  and
      Post, Matt  and
      Specia, Lucia  and
      Turchi, Marco  and
      Verspoor, Karin",
    booktitle = "Proceedings of the Third Conference on Machine Translation: Research Papers",
    month = oct,
    year = "2018",
    address = "Brussels, Belgium",
    publisher = "Association for Computational Linguistics",
    url = "https://aclanthology.org/W18-6319/",
    doi = "10.18653/v1/W18-6319",
    pages = "186--191"
}

@misc{grattafiori2024llama3herdmodels,
      title={The Llama 3 Herd of Models}, 
      author={Aaron Grattafiori and Abhimanyu Dubey and Abhinav Jauhri and Abhinav Pandey and Abhishek Kadian and Ahmad Al-Dahle and Aiesha Letman and Akhil Mathur and Alan Schelten and Alex Vaughan and Amy Yang and Angela Fan and Anirudh Goyal and Anthony Hartshorn and Aobo Yang and Archi Mitra and Archie Sravankumar and Artem Korenev and Arthur Hinsvark and Arun Rao and Aston Zhang and Aurelien Rodriguez and Austen Gregerson and Ava Spataru and Baptiste Roziere and Bethany Biron and Binh Tang and Bobbie Chern and Charlotte Caucheteux and Chaya Nayak and Chloe Bi and Chris Marra and Chris McConnell and Christian Keller and Christophe Touret and Chunyang Wu and Corinne Wong and Cristian Canton Ferrer and Cyrus Nikolaidis and Damien Allonsius and Daniel Song and Danielle Pintz and Danny Livshits and Danny Wyatt and David Esiobu and Dhruv Choudhary and Dhruv Mahajan and Diego Garcia-Olano and Diego Perino and Dieuwke Hupkes and Egor Lakomkin and Ehab AlBadawy and Elina Lobanova and Emily Dinan and Eric Michael Smith and Filip Radenovic and Francisco Guzmán and Frank Zhang and Gabriel Synnaeve and Gabrielle Lee and Georgia Lewis Anderson and Govind Thattai and Graeme Nail and Gregoire Mialon and Guan Pang and Guillem Cucurell and Hailey Nguyen and Hannah Korevaar and Hu Xu and Hugo Touvron and Iliyan Zarov and Imanol Arrieta Ibarra and Isabel Kloumann and Ishan Misra and Ivan Evtimov and Jack Zhang and Jade Copet and Jaewon Lee and Jan Geffert and Jana Vranes and Jason Park and Jay Mahadeokar and Jeet Shah and Jelmer van der Linde and Jennifer Billock and Jenny Hong and Jenya Lee and Jeremy Fu and Jianfeng Chi and Jianyu Huang and Jiawen Liu and Jie Wang and Jiecao Yu and Joanna Bitton and Joe Spisak and Jongsoo Park and Joseph Rocca and Joshua Johnstun and Joshua Saxe and Junteng Jia and Kalyan Vasuden Alwala and Karthik Prasad and Kartikeya Upasani and Kate Plawiak and Ke Li and Kenneth Heafield and Kevin Stone and Khalid El-Arini and Krithika Iyer and Kshitiz Malik and Kuenley Chiu and Kunal Bhalla and Kushal Lakhotia and Lauren Rantala-Yeary and Laurens van der Maaten and Lawrence Chen and Liang Tan and Liz Jenkins and Louis Martin and Lovish Madaan and Lubo Malo and Lukas Blecher and Lukas Landzaat and Luke de Oliveira and Madeline Muzzi and Mahesh Pasupuleti and Mannat Singh and Manohar Paluri and Marcin Kardas and Maria Tsimpoukelli and Mathew Oldham and Mathieu Rita and Maya Pavlova and Melanie Kambadur and Mike Lewis and Min Si and Mitesh Kumar Singh and Mona Hassan and Naman Goyal and Narjes Torabi and Nikolay Bashlykov and Nikolay Bogoychev and Niladri Chatterji and Ning Zhang and Olivier Duchenne and Onur Çelebi and Patrick Alrassy and Pengchuan Zhang and Pengwei Li and Petar Vasic and Peter Weng and Prajjwal Bhargava and Pratik Dubal and Praveen Krishnan and Punit Singh Koura and Puxin Xu and Qing He and Qingxiao Dong and Ragavan Srinivasan and Raj Ganapathy and Ramon Calderer and Ricardo Silveira Cabral and Robert Stojnic and Roberta Raileanu and Rohan Maheswari and Rohit Girdhar and Rohit Patel and Romain Sauvestre and Ronnie Polidoro and Roshan Sumbaly and Ross Taylor and Ruan Silva and Rui Hou and Rui Wang and Saghar Hosseini and Sahana Chennabasappa and Sanjay Singh and Sean Bell and Seohyun Sonia Kim and Sergey Edunov and Shaoliang Nie and Sharan Narang and Sharath Raparthy and Sheng Shen and Shengye Wan and Shruti Bhosale and Shun Zhang and Simon Vandenhende and Soumya Batra and Spencer Whitman and Sten Sootla and Stephane Collot and Suchin Gururangan and Sydney Borodinsky and Tamar Herman and Tara Fowler and Tarek Sheasha and Thomas Georgiou and Thomas Scialom and Tobias Speckbacher and Todor Mihaylov and Tong Xiao and Ujjwal Karn and Vedanuj Goswami and Vibhor Gupta and Vignesh Ramanathan and Viktor Kerkez and Vincent Gonguet and Virginie Do and Vish Vogeti and Vítor Albiero and Vladan Petrovic and Weiwei Chu and Wenhan Xiong and Wenyin Fu and Whitney Meers and Xavier Martinet and Xiaodong Wang and Xiaofang Wang and Xiaoqing Ellen Tan and Xide Xia and Xinfeng Xie and Xuchao Jia and Xuewei Wang and Yaelle Goldschlag and Yashesh Gaur and Yasmine Babaei and Yi Wen and Yiwen Song and Yuchen Zhang and Yue Li and Yuning Mao and Zacharie Delpierre Coudert and Zheng Yan and Zhengxing Chen and Zoe Papakipos and Aaditya Singh and Aayushi Srivastava and Abha Jain and Adam Kelsey and Adam Shajnfeld and Adithya Gangidi and Adolfo Victoria and Ahuva Goldstand and Ajay Menon and Ajay Sharma and Alex Boesenberg and Alexei Baevski and Allie Feinstein and Amanda Kallet and Amit Sangani and Amos Teo and Anam Yunus and Andrei Lupu and Andres Alvarado and Andrew Caples and Andrew Gu and Andrew Ho and Andrew Poulton and Andrew Ryan and Ankit Ramchandani and Annie Dong and Annie Franco and Anuj Goyal and Aparajita Saraf and Arkabandhu Chowdhury and Ashley Gabriel and Ashwin Bharambe and Assaf Eisenman and Azadeh Yazdan and Beau James and Ben Maurer and Benjamin Leonhardi and Bernie Huang and Beth Loyd and Beto De Paola and Bhargavi Paranjape and Bing Liu and Bo Wu and Boyu Ni and Braden Hancock and Bram Wasti and Brandon Spence and Brani Stojkovic and Brian Gamido and Britt Montalvo and Carl Parker and Carly Burton and Catalina Mejia and Ce Liu and Changhan Wang and Changkyu Kim and Chao Zhou and Chester Hu and Ching-Hsiang Chu and Chris Cai and Chris Tindal and Christoph Feichtenhofer and Cynthia Gao and Damon Civin and Dana Beaty and Daniel Kreymer and Daniel Li and David Adkins and David Xu and Davide Testuggine and Delia David and Devi Parikh and Diana Liskovich and Didem Foss and Dingkang Wang and Duc Le and Dustin Holland and Edward Dowling and Eissa Jamil and Elaine Montgomery and Eleonora Presani and Emily Hahn and Emily Wood and Eric-Tuan Le and Erik Brinkman and Esteban Arcaute and Evan Dunbar and Evan Smothers and Fei Sun and Felix Kreuk and Feng Tian and Filippos Kokkinos and Firat Ozgenel and Francesco Caggioni and Frank Kanayet and Frank Seide and Gabriela Medina Florez and Gabriella Schwarz and Gada Badeer and Georgia Swee and Gil Halpern and Grant Herman and Grigory Sizov and Guangyi and Zhang and Guna Lakshminarayanan and Hakan Inan and Hamid Shojanazeri and Han Zou and Hannah Wang and Hanwen Zha and Haroun Habeeb and Harrison Rudolph and Helen Suk and Henry Aspegren and Hunter Goldman and Hongyuan Zhan and Ibrahim Damlaj and Igor Molybog and Igor Tufanov and Ilias Leontiadis and Irina-Elena Veliche and Itai Gat and Jake Weissman and James Geboski and James Kohli and Janice Lam and Japhet Asher and Jean-Baptiste Gaya and Jeff Marcus and Jeff Tang and Jennifer Chan and Jenny Zhen and Jeremy Reizenstein and Jeremy Teboul and Jessica Zhong and Jian Jin and Jingyi Yang and Joe Cummings and Jon Carvill and Jon Shepard and Jonathan McPhie and Jonathan Torres and Josh Ginsburg and Junjie Wang and Kai Wu and Kam Hou U and Karan Saxena and Kartikay Khandelwal and Katayoun Zand and Kathy Matosich and Kaushik Veeraraghavan and Kelly Michelena and Keqian Li and Kiran Jagadeesh and Kun Huang and Kunal Chawla and Kyle Huang and Lailin Chen and Lakshya Garg and Lavender A and Leandro Silva and Lee Bell and Lei Zhang and Liangpeng Guo and Licheng Yu and Liron Moshkovich and Luca Wehrstedt and Madian Khabsa and Manav Avalani and Manish Bhatt and Martynas Mankus and Matan Hasson and Matthew Lennie and Matthias Reso and Maxim Groshev and Maxim Naumov and Maya Lathi and Meghan Keneally and Miao Liu and Michael L. Seltzer and Michal Valko and Michelle Restrepo and Mihir Patel and Mik Vyatskov and Mikayel Samvelyan and Mike Clark and Mike Macey and Mike Wang and Miquel Jubert Hermoso and Mo Metanat and Mohammad Rastegari and Munish Bansal and Nandhini Santhanam and Natascha Parks and Natasha White and Navyata Bawa and Nayan Singhal and Nick Egebo and Nicolas Usunier and Nikhil Mehta and Nikolay Pavlovich Laptev and Ning Dong and Norman Cheng and Oleg Chernoguz and Olivia Hart and Omkar Salpekar and Ozlem Kalinli and Parkin Kent and Parth Parekh and Paul Saab and Pavan Balaji and Pedro Rittner and Philip Bontrager and Pierre Roux and Piotr Dollar and Polina Zvyagina and Prashant Ratanchandani and Pritish Yuvraj and Qian Liang and Rachad Alao and Rachel Rodriguez and Rafi Ayub and Raghotham Murthy and Raghu Nayani and Rahul Mitra and Rangaprabhu Parthasarathy and Raymond Li and Rebekkah Hogan and Robin Battey and Rocky Wang and Russ Howes and Ruty Rinott and Sachin Mehta and Sachin Siby and Sai Jayesh Bondu and Samyak Datta and Sara Chugh and Sara Hunt and Sargun Dhillon and Sasha Sidorov and Satadru Pan and Saurabh Mahajan and Saurabh Verma and Seiji Yamamoto and Sharadh Ramaswamy and Shaun Lindsay and Shaun Lindsay and Sheng Feng and Shenghao Lin and Shengxin Cindy Zha and Shishir Patil and Shiva Shankar and Shuqiang Zhang and Shuqiang Zhang and Sinong Wang and Sneha Agarwal and Soji Sajuyigbe and Soumith Chintala and Stephanie Max and Stephen Chen and Steve Kehoe and Steve Satterfield and Sudarshan Govindaprasad and Sumit Gupta and Summer Deng and Sungmin Cho and Sunny Virk and Suraj Subramanian and Sy Choudhury and Sydney Goldman and Tal Remez and Tamar Glaser and Tamara Best and Thilo Koehler and Thomas Robinson and Tianhe Li and Tianjun Zhang and Tim Matthews and Timothy Chou and Tzook Shaked and Varun Vontimitta and Victoria Ajayi and Victoria Montanez and Vijai Mohan and Vinay Satish Kumar and Vishal Mangla and Vlad Ionescu and Vlad Poenaru and Vlad Tiberiu Mihailescu and Vladimir Ivanov and Wei Li and Wenchen Wang and Wenwen Jiang and Wes Bouaziz and Will Constable and Xiaocheng Tang and Xiaojian Wu and Xiaolan Wang and Xilun Wu and Xinbo Gao and Yaniv Kleinman and Yanjun Chen and Ye Hu and Ye Jia and Ye Qi and Yenda Li and Yilin Zhang and Ying Zhang and Yossi Adi and Youngjin Nam and Yu and Wang and Yu Zhao and Yuchen Hao and Yundi Qian and Yunlu Li and Yuzi He and Zach Rait and Zachary DeVito and Zef Rosnbrick and Zhaoduo Wen and Zhenyu Yang and Zhiwei Zhao and Zhiyu Ma},
      year={2024},
      eprint={2407.21783},
      archivePrefix={arXiv},
      primaryClass={cs.AI},
      url={https://arxiv.org/abs/2407.21783}, 
}

@inproceedings{2024ayamodelinstructionfinetuned,
    title = "Aya Model: An Instruction Finetuned Open-Access Multilingual Language Model",
    author = {{\"U}st{\"u}n, Ahmet  and
      Aryabumi, Viraat  and
      Yong, Zheng  and
      Ko, Wei-Yin  and
      D{'}souza, Daniel  and
      Onilude, Gbemileke  and
      Bhandari, Neel  and
      Singh, Shivalika  and
      Ooi, Hui-Lee  and
      Kayid, Amr  and
      Vargus, Freddie  and
      Blunsom, Phil  and
      Longpre, Shayne  and
      Muennighoff, Niklas  and
      Fadaee, Marzieh  and
      Kreutzer, Julia  and
      Hooker, Sara},
    editor = "Ku, Lun-Wei  and
      Martins, Andre  and
      Srikumar, Vivek",
    booktitle = "Proceedings of the 62nd Annual Meeting of the Association for Computational Linguistics (Volume 1: Long Papers)",
    month = aug,
    year = "2024",
    address = "Bangkok, Thailand",
    publisher = "Association for Computational Linguistics",
    url = "https://aclanthology.org/2024.acl-long.845/",
    doi = "10.18653/v1/2024.acl-long.845",
    pages = "15894--15939"
}

@inproceedings{liu-etal-2021-usefulness,
    title = "The Usefulness of {B}ibles in Low-Resource Machine Translation",
    author = "Liu, Ling  and
      Ryan, Zach  and
      Hulden, Mans",
    editor = "Arppe, Antti  and
      Good, Jeff  and
      Harrigan, Atticus  and
      Hulden, Mans  and
      Lachler, Jordan  and
      Moeller, Sarah  and
      Palmer, Alexis  and
      Silfverberg, Miikka  and
      Schwartz, Lane",
    booktitle = "Proceedings of the 4th Workshop on the Use of Computational Methods in the Study of Endangered Languages Volume 1 (Papers)",
    month = mar,
    year = "2021",
    address = "Online",
    publisher = "Association for Computational Linguistics",
    url = "https://aclanthology.org/2021.computel-1.6/",
    pages = "44--50"
}

@misc{Claude3S,
  title={Claude 3.5 Sonnet Model Card Addendum},
  author={Anthropic},
  year={2020},
  url={https://api.semanticscholar.org/CorpusID:270667923}
}

@inproceedings{zhu-etal-2024-multilingual,
    title = "Multilingual Machine Translation with Large Language Models: Empirical Results and Analysis",
    author = "Zhu, Wenhao  and
      Liu, Hongyi  and
      Dong, Qingxiu  and
      Xu, Jingjing  and
      Huang, Shujian  and
      Kong, Lingpeng  and
      Chen, Jiajun  and
      Li, Lei",
    editor = "Duh, Kevin  and
      Gomez, Helena  and
      Bethard, Steven",
    booktitle = "Findings of the Association for Computational Linguistics: NAACL 2024",
    month = jun,
    year = "2024",
    address = "Mexico City, Mexico",
    publisher = "Association for Computational Linguistics",
    url = "https://aclanthology.org/2024.findings-naacl.176/",
    doi = "10.18653/v1/2024.findings-naacl.176",
    pages = "2765--2781"
}

@misc{PavaEtAl2025MindGap,
  title       = {Mind the (Language) Gap: Mapping the Challenges of {LLM} Development in Low-Resource Language Contexts},
  author      = {Pava, Juan N. and Meinhardt, Caroline and Badi Uz Zaman, Haifa and Friedman, Toni and Truong, Sang T. and Zhang, Daniel and Cryst, Elena and Marivate, Vukosi and Koyejo, Sanmi},
  year        = {2025},
  institution = {HAI},
  type        = {White Paper},
  url = {https://hai.stanford.edu/policy/mind-the-language-gap-mapping-the-challenges-of-llm-development-in-low-resource-language-contexts}
}

@misc{frontull2025compensatingdatareasoninglowresource,
      title={Compensating for Data with Reasoning: Low-Resource Machine Translation with {LLM}s}, 
      author={Samuel Frontull and Thomas Ströhle},
      year={2025},
      eprint={2505.22293},
      archivePrefix={arXiv},
      primaryClass={cs.CL},
      url={https://arxiv.org/abs/2505.22293}, 
}

@inproceedings{zeldes-etal-2025-ud,
    title = "A {UD} Treebank for Bohairic {C}optic",
    author = "Zeldes, Amir  and
      Speransky, Nina  and
      Wagner, Nicholas E.  and
      Schroeder, Caroline T.",
    editor = {Bouma, Gosse  and
      {\c{C}}{\"o}ltekin, {\c{C}}a{\u{g}}r{\i}},
    booktitle = "Proceedings of the Eighth Workshop on Universal Dependencies (UDW, SyntaxFest 2025)",
    month = aug,
    year = "2025",
    address = "Ljubljana, Slovenia",
    publisher = "Association for Computational Linguistics",
    url = "https://aclanthology.org/2025.udw-1.7/",
    pages = "59--69",
    ISBN = "979-8-89176-292-3"
}

@inproceedings{slaughter-etal-2019-making,
    title = "The Making of {C}optic {W}ordnet",
    author = {Slaughter, Laura  and
      Costa, Luis Morgado Da  and
      Miyagawa, So  and
      B{\"u}chler, Marco  and
      Zeldes, Amir  and
      Behlmer, Heike},
    editor = "Vossen, Piek  and
      Fellbaum, Christiane",
    booktitle = "Proceedings of the 10th Global Wordnet Conference",
    month = jul,
    year = "2019",
    address = "Wroclaw, Poland",
    publisher = "Global Wordnet Association",
    url = "https://aclanthology.org/2019.gwc-1.21/",
    doi = "10.18653/v1/2019.gwc-1.21",
    pages = "166--175"
}

@inproceedings{PetrovDasMcDonald2012,
    title = "A Universal Part-of-Speech Tagset",
    author = "Petrov, Slav  and
      Das, Dipanjan  and
      McDonald, Ryan",
    editor = "Calzolari, Nicoletta  and
      Choukri, Khalid  and
      Declerck, Thierry  and
      Do{\u{g}}an, Mehmet U{\u{g}}ur  and
      Maegaard, Bente  and
      Mariani, Joseph  and
      Moreno, Asuncion  and
      Odijk, Jan  and
      Piperidis, Stelios",
    booktitle = "Proceedings of the Eighth International Conference on Language Resources and Evaluation ({LREC}'12)",
    month = may,
    year = "2012",
    address = "Istanbul, Turkey",
    publisher = "European Language Resources Association (ELRA)",
    url = "https://aclanthology.org/L12-1115/",
    pages = "2089--2096"
}

@Article{ZeldesSchroeder2015,
  author    = {Amir Zeldes and Caroline T. Schroeder},
  title     = {Computational Methods for {C}optic: {D}eveloping and Using Part-of-Speech Tagging for Digital Scholarship in the Humanities},
  journal   = {Digital Scholarship in the Humanities},
  year      = {2015},
  volume    = {30},
  number    = {1},
  pages     = {164--176},
  doi = {10.1093/llc/fqv043},
  url = {https://academic.oup.com/dsh/article-abstract/30/suppl_1/i164/364067}
}

@article{vangysel2021umr,
	title        = {Designing a Uniform Meaning Representation for Natural Language Processing},
	author       = {Van Gysel, Jens and Vigus, Meagan and Chun, Jayeol and Lai, Kenneth and Moeller, Sarah and Yao, Jiarui and O'Gorman, Timothy J. and Cowell, Andrew and Croft, W. Bruce and Huang, Chu-Ren and Haji\v{c}, Jan and Martin, James H. and Oepen, Stephan and Palmer, Martha and Pustejovsky, James and Vallejos, Rosa and Xue, Nianwen},
	year         = 2021,
	journal      = {K{\"u}nstliche Intelligenz},
	volume       = 35,
	pages        = {343--360},
	doi          = {10.1007/s13218-021-00722-w},
	url          = {https://link.springer.com/article/10.1007/s13218-021-00722-w}
}

@article{haddow-etal-2022-survey,
    title = "Survey of Low-Resource Machine Translation",
    author = "Haddow, Barry  and
      Bawden, Rachel  and
      Miceli Barone, Antonio Valerio  and
      Helcl, Jind{\v{r}}ich  and
      Birch, Alexandra",
    journal = "Computational Linguistics",
    volume = "48",
    number = "3",
    month = sep,
    year = "2022",
    address = "Cambridge, MA",
    publisher = "MIT Press",
    url = "https://aclanthology.org/2022.cl-3.6/",
    doi = "10.1162/coli_a_00446",
    pages = "673--732"
}

@inproceedings{hus-anastasopoulos-2024-back,
    title = "Back to School: Translation Using Grammar Books",
    author = "Hus, Jonathan  and
      Anastasopoulos, Antonios",
    editor = "Al-Onaizan, Yaser  and
      Bansal, Mohit  and
      Chen, Yun-Nung",
    booktitle = "Proceedings of the 2024 Conference on Empirical Methods in Natural Language Processing",
    month = nov,
    year = "2024",
    address = "Miami, Florida, USA",
    publisher = "Association for Computational Linguistics",
    url = "https://aclanthology.org/2024.emnlp-main.1127/",
    doi = "10.18653/v1/2024.emnlp-main.1127",
    pages = "20207--20219"
}

@misc{burns2020coptic_lexicon,
	title        = {Comprehensive {C}optic Lexicon: Including Loanwords from {A}ncient {G}reek},
	author       = {Burns, Dylan M. and Feder, Frank and John, Katrin and Kupreyev, Maxim},
	year         = 2020,
	doi          = {10.17169/refubium-27566},
	url          = {https://refubium.fu-berlin.de/handle/fub188/27813},
	version      = {1.2}
}

@misc{bapna2022building,
	title        = {Building machine translation systems for the next thousand languages},
	author       = {Bapna, Ankur and Caswell, Isaac and Kreutzer, Julia and Firat, Orhan and van Esch, Daan and Siddhant, Aditya and Niu, Mengmeng and Baljekar, Pallavi and Garcia, Xavier and Macherey, Wolfgang and others},
	year         = 2022,
	journal      = {arXiv preprint arXiv:2205.03983},
	doi          = {https://doi.org/10.48550/arXiv.2205.03983},
	url          = {https://arxiv.org/abs/2205.03983}
}

@article{almond2013DDGLC,
  title={Kontaktinduzierter Sprachwandel des {\"A}gyptisch-Koptischen: Lehnwort-Lexikographie im Projekt Database and Dictionary of Greek Loanwords in Coptic (DDGLC)},
  author={Almond, Mathew and Hagen, Joost and John, Katrin and Richter, Tonio Sebastian and Walter, Vincent},
  journal={Perspektiven einer corpusbasierten historischen Linguistik und Philologie},
  pages={283--315},
  year={2013}
}

@book{ethnologue_2026,
	address = {Dallas, Texas},
	title = {Ethnologue: {Languages} of the {World}. {Twenty}-ninth edition},
	url = {http://www.ethnologue.com.},
	publisher = {SIL International},
	editor = {Eberhard, David M. and Simons, Gary F. and Robinson, Alison J.},
	year = {2026},
}

\appendix
\section{Data and Resources}

\begin{table}[htb!]
    \small
    \centering
    \begin{tabular}{l l}
        \textbf{Resource} & \textbf{Reference}\\
        \toprule
         UD Treebank &  \citealp{zeldes-abrams-2018-coptic}\\
         Ostraca & \citealp{wannaz-miyagawa-2024-assessing}\\
         
         Coptic Dictionary & \citealp{feder-etal-2018-linked}\\
         Coptic NLP & \href{https://copticscriptorium.org/}{Coptic Scriptorium} \\
         \bottomrule
    \end{tabular}
    \caption{The different data and resources used in this paper.}
    \label{tab:resources}
\end{table}

\Cref{tab:resources} collates the core resources used in this paper.

\subsection*{Details about the Dictionary}
The lexicon contains multiple entries for a single form.

\ex. \begin{coptic}eimhti\end{coptic}\footnote{\url{https://coptic-dictionary.org/results.py?quick\_search=C8880}} 
\a.  except
\b. nevertheless

A specific entry can contain elaborate information, including translations in multiple languages, and for multiple dialects of Coptic. We focus on and use the Sahidic dialect.

\ex. except\footnote{\url{https://coptic-dictionary.org/entry.py?tla=C8880}}
\a. except (for), if not
\b. except (for), without that, unless
\b. but

\section{System and Experiments}
\label{sec:appendix-sys-and-experiments}
\subsection{Discussion on BLEU and \chrf}
\label{sec:appendix-metrics}
We also report \chrf~\cite{popovic-2017-chrf} for many of the experiments as an additional reference metric.

We note that, as discussed in \citet{post-2018-call}, BLEU can be parameterized in multiple ways. We use the implementation from the \texttt{sacrebleu} package\footnote{\url{https://github.com/mjpost/sacrebleu}}, and use default settings for \dev~and ostraca\footnote{The \texttt{sacrebleu} metric signature is \path{nrefs:379|case:mixed|eff:no|tok:13a|smooth:none|version:2.5.1}}. Since BLEU is designed as a corpus level metric, the default settings of BLEU lead to some $0$ scored outputs for the \test~split, so we used a relaxed setting with max 3-gram, effective order, and `floor` smoothing with $0.1$\footnote{The metric signature is \path{nrefs:405|case:mixed|eff:yes|tok:13a|smooth:floor[0.10]|version:2.5.1}.}.

BLEU is specifically not useful for the \test~set since, although the floor smoothing allows for non-zero BLEU scores, but leads to the common value $13.56$ for many settings. BLEU is not the primary metric in the paper, but is kept for completness.

\paragraph{BLEU is unreliable in \test}
\label{sec:bleu-test-unreliable}
BLEU as a metric was somewhat representative if not correlated with BertScore for \dev~and ostraca, but with default signature the BLEU scores for \test~ collapsed to $0$ for many settings. We have reported a smoothed BLEU for~\test~to be consistent. 

This is mainly a function of the model output utilizing a wider variety of surface forms in conflict with an n-gram overlap mechanism that BLEU is based upon, but is not distinct from size of the reference or the output sentence.

Consider the following translation sentence: `For you have taken the things of the poor and the widows and the orphans, and you have placed them in your window.' Gemma-3-12b with the \dep+\con~setting responds simply with {`You did.'} This leads to a $0$ BLEU score  without smoothing. Overall there are only $29$ of the $405$ sentences that show this behavior across the three test models we report in \cref{tab:test-components}.  This highlights for Coptic the need for further work in validating the representativeness and quality of the wide-ranging metrics that exist for MT~\cite{lavie-etal-2025-findings}.

\paragraph{BLEU is inconsistent in \dev}\label{sec:appendix-dev-bleu}
Unlike with \test, BLEU is not unstable on \dev; however it is inconsistent. In \cref{tab:dev-components}, for Gemma-12b we can see that BLEU does not follow the same pattern as BertScore. This is partly due to the difference in the two class of metrics. Since BLEU relies on n-gram precision, it is more sensitive to lexical choice.

Compare the output translation `And the remainder of the great ones in the palace.' with the reference translation `as did also the other nobles who were attached to the Palace.' The translation did capture some semantic elements such as `great ones' in comparison to `nobles' but by design, BLEU would consider 0-precision, while BertScore would catch some quotient of similarity. Such patterns are more common in LLM based generation, compared to older controlled or limited vocabulary methods.

\label{sec:appendix-components}
% \subsection{Lexicon}
\subsection{\lexicon~parameters and grid search}
\label{sec:appendix-lexgridsearch}
Our \lexicon~component has multiple configurable parameters, one controlling the target languages we include from dictionary and the rest for controlling the information added. 

Two parameters control the first-k entries that we include after retrieval and the first-k senses for each entry that we include. The final parameter is for a deduplication feature to specifically handle the verbose enumeration for the DDGLC~\cite{almond2013DDGLC} portion of the dictionary.

The grid search for \lexicon~was performed with the best setting for \dep~as decided by the previous grid search. Then the \lexicon~grid search was performed on the sub-selected 10 each of the easiest and hardest to translate for our pilot model. 

\paragraph{Fixed setting} The final parameters were to use only entries from English, no more than 100 entries per sample, with no more than 10 senses per entry, and without deduplicating the information from DDGLC.

\subsection{\dep~parameters and grid Search}
\label{sec:appendix-depgridsearch}

Our \dep~component is configurable with three different parameters controlling how duplicate tokens are handled, the robustness of verbalized dependency relations, and the inclusion of various `tiers' of POS tags. 

When handling duplicates, we choose subscript notation ('\begin{coptic}eh1rai\end{coptic}$_2$ is the case marking of \begin{coptic}yoyou\end{coptic}') or verbalized nominalizations ('the second \begin{coptic}eh1rai\end{coptic} is the case marking of \begin{coptic}yoyou\end{coptic}). 

When choosing which dependency relations to verbalize in this component, we either `collapse' the relations or maintain the full list. For example, the non-collapsed setting verbalizes each dependency relation as they are defined in UD: \textit{ccomp} is expressed as a clausal complement, and \textit{xcomp} as an open clausal complement. In the collapsed setting, both \textit{ccomp} and \textit{xcomp} are verbalized as simply `complement'.

Finally, we choose from three tiers of POS tags to verbalize in this \dep~component: content, participants, or all. We excluded punctuation, symbols, and `other' tags across the board. The content tier is the most restrictive, allowing only for nouns, verbs, proper nouns, adpositions, and adverbs. This setting results in the shortest final \dep~component as it filters out the most relations from the final verbalization. The participants tier expands upon the content tier, adding pronouns, auxiliaries, determinants, and numerals. The final tier is the most inclusive, adding conjunctions to the previous list. 

\paragraph{Fixed setting} Following our grid search, we determined that our best setting is to verbalize duplicates as \textit{subscripts, not collapse} the dependency relations into more general categories, and to use the \textit{participants tier} of POS tags. 

\subsection{Constructions \con}
\label{sec:appendix-constructions}

The \con~ component provides information and instruction targeting specific source language constructions. Construction prompts are triggered by 26 DepEdit rules which identify syntax subtrees using a declaration of nodes to be found and subgraph relations which must hold between them. 

For example, when a verb is accompanied by both the future and preterit auxiliaries, whose meanings individually correspond to future tense (`will VERB') and continuous past (`was VERBing'), the combination results in a counterfactual conditional (`would have VERBed'). This is captured by declaring three nodes to match: the VERB (via POS tag) and the two dependents (the auxiliaries, via their lemmas and POS tags) in a particular order (preterite, then future, then the VERB). 

The DepEdit module rule can then extract annotations attached to these nodes, such as their form (since both the VERB and auxiliaries can inflect), lemma, or morphological categories such as person or gender. The template for the counterfactual construction is verbalized as:

\ex. The combination of the future auxiliary \{\{\textsc{fut. aux form}\}\} with the counterfactual preterit \{\{\textsc{pret. aux form}\}\} is used together with the predicate \{\{\textsc{verb form}\}\} to express a counterfactual meaning (would have VERBed).

The list of construction triggers and their verbalizations is available with the full code in the paper's GitHub repository.

\subsection{Instruction Design}

\begin{figure*}[t]
    \textbf{Base}
    \tcbset{colback=white}
    \begin{mybox}\sf
    You are a professional Coptic-to-English translator tasked with providing translations suitable for use in United States (en\_US). Your goal is to accurately convey the meaning and nuances of the original Coptic text while adhering to English grammar, vocabulary, and cultural sensitivities.Produce only the English translation, without any additional explanations or commentary.Please translate the following Coptic text into English (en\_US):  \{\textit{source}\}.
    \end{mybox}
    \label{fig:base-instruction}
    \textbf{Closing cue for consistency.}
    \begin{mybox}\sf
Using all the information provided above, now please translate the sentence into English(en\_US). Remember your source sentence is: {\{\textit{source}\}}.The English translation is: 
\end{mybox}
    \label{fig:consistency-instruction.}
    \textbf{Lexicon}
    \begin{mybox}\sf
    For the translation task, you are given dictionary entries for Coptic. Some words can be polysemous and there might be multiple entries. Each entry can contain multiple senses with translations in [`English']. In such a case, please choose the most appropriate one. Note that for some words, they might be derived from a more basic form, some entries will be for such lemma.

     Here are the entries for collected for individual words in the sentence:\\ \\Dictionary entry Verb \begin{coptic}d1w\end{coptic} has 2 senses.\\
     Sense 1:\\
     - In English, \begin{coptic}d1w\end{coptic} means say, speak, tell\\
     Sense 2:\\
     - In English, \begin{coptic}d1w\end{coptic} means sing
    ...
    \end{mybox}
    \label{fig:lexicon-instruction}
     \textbf{Dependency}
    \begin{mybox}\sf
    The dependency information for the sentence is:
    \begin{coptic}rmh\end{coptic} is the root. \begin{coptic}eh1rai\end{coptic}$_1$ is the case marking of \begin{coptic}baukalion\end{coptic}. \begin{coptic}e\end{coptic}$_1$ is the fixed multiword expression of \begin{coptic}eh1rai\end{coptic}$_1$. \begin{coptic}baukalion\end{coptic} is the oblique nominal of \begin{coptic}rmh\end{coptic}. \begin{coptic}h\end{coptic} is the coordinating conjunction of \begin{coptic}yoyou\end{coptic}. \begin{coptic}eh1rai\end{coptic}$_2$ is the case marking of \begin{coptic}yoyou\end{coptic}. \begin{coptic}e\end{coptic}$_2$ is the fixed multiword expression of \begin{coptic}eh1rai\end{coptic}$_2$. \begin{coptic}yoyou\end{coptic} is the conjunct of \begin{coptic}baukalion\end{coptic}. \begin{coptic}eh1rai\end{coptic}$_3$ is the case marking of \begin{coptic}laau\end{coptic} ...
    \end{mybox}
    
    \textbf{Construction}\sf
    \begin{mybox}\sf
    The information about specific constructions in the sentence is:
        The dislocated element \begin{coptic}pai\end{coptic} is a repeated reference to the pronoun dependent of the predicate \begin{coptic}d1w\end{coptic}. There is often no need to translate the pronominal mention of the same argument. ...
    \end{mybox}
\caption{An example of the content in differrent sections of the Instruction to LM. The CONLL-U is separately shown in \cref{fig:conllu}.}
    \label{fig:instruction-design}
\end{figure*}

\begin{figure*}[hbt!]
    \textbf{CoNLL-U}
    \begin{mybox}\sf
    \begin{tabular}{llllllllll}
   1 & \begin{coptic}m\end{coptic} & \begin{coptic}n\end{coptic} & ADP & PREP & \_ & 3 & case & \_ & \_\\
2 & \begin{coptic}p\end{coptic} & \begin{coptic}p\end{coptic} & DET & ART & \_ & 3 & det & \_ & \_\\
3 & \begin{coptic}h1agioc\end{coptic} & \begin{coptic}h1agios\end{coptic} & NOUN & N & \_ & 0 & root & \_ & \_\\
4 & \begin{coptic}biktwr\end{coptic} & \begin{coptic}biktwr\end{coptic} & PROPN & NPROP & \_ & 3 & appos & \_ & \_\\
5 & \begin{coptic}pe\end{coptic} & \begin{coptic}p\end{coptic} & DET & ART & \_ & 6 & det & \_ & \_\\
6 & \begin{coptic}crathlathc\end{coptic} & \begin{coptic}ctrathlathc\end{coptic} & NOUN & N & \_ & 3 & appos & \_ & \_\\
7 & \begin{coptic}auw\end{coptic} & \begin{coptic}auw\end{coptic} & CCONJ & CONJ & \_ & 9 & cc & \_ & \_\\
8 & \begin{coptic}p\end{coptic} & \begin{coptic}p\end{coptic} & DET & ART & \_ & 9 & det & \_ & \_\\
9 & \begin{coptic}marturoc\end{coptic} & \begin{coptic}marturoc\end{coptic} & NOUN & N & \_ & 6 & conj & \_ & \_\\
10 & \begin{coptic}et\end{coptic} & \begin{coptic}etere\end{coptic} & SCONJ & CREL & \_ & 11 & mark & \_ & \_\\
11 & \begin{coptic}taihu\end{coptic} & \begin{coptic}taeio\end{coptic} & VERB & VSTAT & \_ & 9 & acl:relcl & \_ & \_\\
12 & \begin{coptic}m\end{coptic} & \begin{coptic}n\end{coptic} & ADP & PREP & \_ & 14 & case & \_ & \_ \\
... & & & & & & & & & 
    \end{tabular}
    \end{mybox}
    \caption{An example CoNLL-U data format, which would also be included into the instruction.}
    \label{fig:conllu}
\end{figure*}

\label{sec:appendix-instruction-design}
\Cref{fig:instruction-design} shows how different parts of the instruction appear. \Cref{fig:conllu} shows the CoNLL-U separate from the rest of the instructions. \cref{fig:instruction-example} shows how these are ordered and provided in our prompt in a condensed manner.

\section{Full results}

\begin{table}[htb!]
    \centering
    \begin{tabular}{ll}
         \textbf{Model (Size)} & \textbf{Model Card}\\
         \toprule
         Gemma3(-it) (12B) &  HF Hub : \href{https://huggingface.co/google/gemma-3-12b-it}{link}\\
         Gemma3(-it) (27B) &   HF Hub : \href{https://huggingface.co/google/gemma-3-27b-it}{link}\\
         GPT-4.1 (NA) & OpenAI Platform : \href{https://platform.openai.com/docs/models/gpt-4.1}{link} \\
         GPT-4.1 Mini (NA) & OpenAI Platform : \href{https://platform.openai.com/docs/models/gpt-4.1-mini}{link} \\
          Llama3.1(-Inst) (8B) &  HF Hub : \href{https://huggingface.co/meta-llama/Llama-3.1-8B-Instruct}{link}\\
         Aya-Expanse (8B) &  HF Hub : \href{https://huggingface.co/CohereLabs/aya-expanse-8b/tree/main}{link}\\
         Aya-Expanse (32B) &  HF Hub : \href{https://huggingface.co/CohereLabs/aya-expanse-32b}{link}\\
         \bottomrule
    \end{tabular}
    \caption{Models used for our experiments with \dev~data and corresponding model cards.}
    \label{tab:model-cards}
\vspace{-10pt}
\end{table}

\label{sec:appendix-full-results}
We reported the results of the Gemma models and GPT-4.1 for \dev~and \test~ in the main body. \Cref{tab:dev-all-results} contains the \dev~results for all the open models we experimented with.
The \dep+\con~setting is reported only for the Gemma models, and not for Aya or Llama models.

Settings using \conll~and \lexicon~exceed the context window of models such as Aya-Expanse, which have smaller context windows (8K compared to the 128K of the other models we used), nonetheless we report it for completeness. This is visible in their performance under those settings, since much of the queue is truncated.

\begin{table*}[t]
    \centering
    \begin{tabular}{llrrr}
        \toprule
        \textbf{Model} & \textbf{Setting} & \textbf{BLEU} & \textbf{BertScore} & \textbf{chrF++} \\
        \midrule
        Aya-Expanse-32b & Baseline	& 3.95& 0.7939 & 14.78\\
        Aya-Expanse-32b & \lexicon & 8.60 & 0.8085 & 17.26 \\
        Aya-Expanse-32b & \conll & 8.25 & 0.7841 & 14.35 \\
        Aya-Expanse-32b & \con & 4.42 & 0.8017 & 15.84 \\
        Aya-Expanse-32b & \dep & 1.63 & 0.7997 & 15.06 \\
        Aya-Expanse-32b & \all & 9.48 & 0.8303 & 22.30 \\[3pt]
        
        Aya-Expanse-8b & Baseline & 6.99 & 0.8132 & 15.34 \\
        Aya-Expanse-8b & \lexicon & 4.40 & 0.8404 & 21.60 \\
        Aya-Expanse-8b & \conll & 2.85 & 0.8265 & 16.51 \\
        Aya-Expanse-8b & \con & 8.16 & 0.8250 & 15.89 \\
        Aya-Expanse-8b & \dep & 6.00 & 0.8073 & 15.11 \\
        Aya-Expanse-8b & \all & 2.39 & 0.8239 & 19.65 \\[3pt]
        
        Gemma-3-12b & Baseline & 16.45 & 0.8342 & 16.29 \\
        Gemma-3-12b & \lexicon & 5.46 & 0.8593 & 23.19 \\
        Gemma-3-12b & \conll & 27.49 & 0.8453 & 18.17 \\
        Gemma-3-12b & \con & 20.26 & 0.8499 & 17.30 \\
        Gemma-3-12b & \dep & 28.66 & 0.8405 & 16.34 \\
        Gemma-3-12b & \dep+\con & 21.65 & 0.8490 & 18.15 \\
        Gemma-3-12b & \all & 7.71 & 0.8722 & 27.29 \\[3pt]
        
        Gemma-3-27b & Baseline & 10.30 & 0.8375 & 17.88 \\
        Gemma-3-27b & \lexicon & 14.46 & 0.8611 & 23.52 \\
        Gemma-3-27b & \conll & 14.86 & 0.8522 & 21.65 \\
        Gemma-3-27b & \con & 15.43 & 0.8509 & 19.27 \\
        Gemma-3-27b & \dep & 15.28 & 0.8414 & 18.83 \\
        Gemma-3-27b & \dep+\con & 10.38 & 0.8509 & 20.05 \\
        Gemma-3-27b & \all & 18.84 & 0.8736 & 28.67 \\[3pt]
        
        Llama-3.1-8B-Instruct & Baseline & 8.42 & 0.7843 & 12.55 \\
        Llama-3.1-8B-Instruct & \lexicon & 3.59 & 0.8004 & 16.30 \\
        Llama-3.1-8B-Instruct & \conll & 6.63 & 0.8009 & 13.30 \\
        Llama-3.1-8B-Instruct & \con & 4.58 & 0.7965 & 14.16 \\
        Llama-3.1-8B-Instruct & \dep & 3.60 & 0.7885 & 12.06 \\
        Llama-3.1-8B-Instruct & \all & 1.94 & 0.8108 & 19.36 \\
        \bottomrule
    \end{tabular}
    \caption{The results from \dev~for all the open-weight models we considered. Note \dep+\con~was run only reported for Gemma models for analysis focused on disambigation.}
    \label{tab:dev-all-results}
\end{table*}

\paragraph{Ostraca}
\Cref{tab:ostraca-result} shows the performance of our Gemma and GPT models across the various settings for the ostraca set.

\begin{table*}[t]
    \centering
    \begin{tabular}{lllll}
        \toprule
        \textbf{Model} & \textbf{Setting} & \textbf{BLEU} & \textbf{BertScore} & \textbf{METEOR} \\
        \midrule
        Gemma-3-12b & Baseline & 5.72 & 0.8380 & 0.11 \\
        Gemma-3-12b & \all & 7.82 & 0.8850 & 0.31 \\
        Gemma-3-12b & \conll & 4.88 & 0.8536 & 0.18 \\
        Gemma-3-12b & \con & 3.84 & 0.8520 & 0.10 \\
        Gemma-3-12b & \dep & 7.50 & 0.8470 & 0.11 \\
        Gemma-3-12b & \lexicon & 5.48 & 0.8640 & 0.26 \\
        Gemma-3-12b & \dep+\con & 6.42 & 0.8591 & 0.12 \\[3pt]
        Gemma-3-27b & Baseline & 5.61 & 0.8490 & 0.13 \\
        Gemma-3-27b & \all & 16.18 & 0.8781 & 0.33 \\
        Gemma-3-27b & \conll & 15.22 & 0.8622 & 0.21 \\
        Gemma-3-27b & \con & 13.32 & 0.8550 & 0.13 \\
        Gemma-3-27b & \dep & 2.39 & 0.8525 & 0.13 \\
        Gemma-3-27b & \lexicon & 9.44 & 0.8621 & 0.22 \\
        Gemma-3-27b & \dep+\con & 9.69 & 0.8599 & 0.16 \\[3pt]
        GPT-4.1 & Baseline & 16.25 & 0.8792 & 0.31 \\
        GPT-4.1 & \all & 17.99 & 0.90 & \textbf{0.53}\\
        GPT-4.1 & \conll & 6.18 & 0.8877 & 0.39 \\
        GPT-4.1 & \con & 9.42 & 0.8888 & 0.31 \\
        GPT-4.1 & \dep & \underline{18.15} & 0.8859 & 0.37 \\
        GPT-4.1 & \lexicon & 14.27 & 0.8973 & 0.42 \\
        GPT-4.1 & \dep+\con & 9.21 & 0.8862 & 0.32 \\[3pt]
        
        \textbf{From \citet{wannaz-miyagawa-2024-assessing}}\\[3pt]
        Claude Opus &  &20.02 & - & 0.46\\
       Claude Haiku &  & 11.52 & - & 0.35\\
        CopticTranslator &  & 8.43 & - & 0.30\\
        \toprule
        \end{tabular}
    \caption{Results for ostraca~\cref{sec:data-ostraca} for Gemma models and GPT-4.1. The BLEU and METEOR scores for Claude (Opus and Haiku) and CopticTranslator are from \citet{wannaz-miyagawa-2024-assessing}.}
    \label{tab:ostraca-result}
\end{table*}

\paragraph{Automatic parses vs gold parses}

\Cref{tab:dev-gold} shows the results for the default usage of automatic parses versus the use of gold parses. Note that our baseline does not use parses in any form, hence performance is equal.

\begin{table*}[t]
    \centering
    \resizebox{\textwidth}{!}{
    \begin{tabular}{llrrrrrr}
         &  & \multicolumn{3}{c}{\textbf{Automatic}} & \multicolumn{3}{c}{\textbf{Gold}} \\
        \toprule
        \textbf{Model} & \textbf{Setting} & \textbf{BLEU} & \textbf{BertScore} & \textbf{chrF++} & \textbf{BLEU} & \textbf{BertScore} & \textbf{chrF++} \\
        \midrule
        Gemma-3-12b & Baseline & 16.45 & 0.8342 & 16.29 & - & - & - \\
        Gemma-3-12b & \dep+\con & 21.65 & \underline{0.8490} & 18.15 & 14.56 & 0.8483 & 18.09 \\
        Gemma-3-12b & \con & 20.26 & 0.8499 & 17.30 & 21.74 & \underline{0.8502} & 17.37 \\
        Gemma-3-12b & \dep & 28.66 & 0.8405 & 16.34 & 28.66 & \underline{0.8411} & 16.18 \\
        Gemma-3-12b & \conll & 27.49 & 0.8453 & 18.17 & 13.50 & \underline{0.8499} & 19.43 \\
        Gemma-3-12b & \lexicon & 5.46 & \underline{0.8593} & 23.19 & 6.45 & 0.8581 & 22.84 \\
        Gemma-3-12b & \all & 7.71 & 0.8722 & 27.29 & 14.22 & \underline{0.8727} & 27.08 \\[3pt]
        Gemma-3-27b & Baseline & 10.30 & 0.8375 & 17.88 & - & - & - \\
        Gemma-3-27b & \dep+\con & 10.38 & 0.8509 & 20.05 & 20.08 & \underline{0.8518} & 20.15 \\
        Gemma-3-27b & \con & 15.43 & 0.8509 & 19.27 & 12.87 & \underline{0.8512} & 19.30 \\
        Gemma-3-27b & \conll & 14.86 & 0.8522 & 21.65 & 10.90 & \underline{0.8545} & 22.68 \\
        Gemma-3-27b & \lexicon & 14.46 & \underline{0.8611} & 23.52 & 9.56 & 0.8604 & 23.22 \\
        Gemma-3-27b & \dep & 15.28 & \underline{0.8414} & 18.83 & 21.39 & 0.8413 & 18.74 \\
        Gemma-3-27b & \all & 18.84 & 0.8736 & 28.67 & 17.11 & \underline{0.8770} & 29.97 \\
        \bottomrule
    \end{tabular}
    }
    \caption{Results for the open-weight Gemma models for \dev~using Automatic and Gold UD parses~(\cref{sec:with-gold}). The higher performance (BertScore) among the two for each setting is underlined. We don't report specific significance for this comparison, although the lower variability across the Automatic and Gold than among the different settings is encouraging.}
    \label{tab:dev-gold}
\end{table*}

\paragraph{Bible Texts}
\begin{table*}[t]
    \centering
    \begin{tabular}{llrrrr}
    & &  \multicolumn{2}{c}{\textbf{Bible}}  &  \multicolumn{2}{c}{\textbf{Other}}  \\
    \toprule
    \textbf{Model} & \textbf{Setting} & \textbf{BLEU} & \textbf{BertScore} & \textbf{BLEU} & \textbf{BertScore}\\
    \midrule
Gemma-3-12b & Baseline & 9.55 & 0.8323 & 9.73 & 0.8297\\
Gemma-3-12b & \all & 19.30 & 0.8738 & 6.17 & 0.8685\\
Gemma-3-12b & \conll & 10.70 & 0.8458 & 18.77 & 0.8436\\
Gemma-3-12b & \con & 17.29 & 0.8537 & 12.55 & 0.8448\\
Gemma-3-12b & \dep & 20.86 & 0.8422 & 18.68 & 0.8373\\
Gemma-3-12b & \lexicon & 26.08 & 0.8585 & 5.03 & 0.8582\\
Gemma-3-12b & \dep+\con & 17.75 & 0.8496 & 20.01 & 0.8467\\[3pt]
Gemma-3-27b & Baseline & 9.78 & 0.8378 & 8.68 & 0.8361\\
Gemma-3-27b & \all & 23.87 & 0.8792 & 14.72 & 0.8665\\
Gemma-3-27b & \conll & 21.87 & 0.8519 & 10.83 & 0.8482\\
Gemma-3-27b & \con & 17.29 & 0.8544 & 13.22 & 0.8453\\
Gemma-3-27b & \dep & 15.46 & 0.8423 & 12.57 & 0.8390\\
Gemma-3-27b & \lexicon & 19.67 & 0.8635 & 13.27 & 0.8577\\
Gemma-3-27b & \dep+\con & 13.13 & 0.8538 & 9.15 & 0.8455\\
\bottomrule
    \end{tabular}
    \caption{Results of Gemma model for \dev~reported as Bible and Not Bible text. The data are distinct but not drastically disproportinate (182 Bible, rest 198 Other).  Performance for Bible data (by BERTScore) is consistently better than Other, which is to interesting, but not suprising for such models w.r.t to low-resource language.}
    \label{tab:dev-bible}
\end{table*}

\Cref{tab:dev-bible} shows the difference in performance between Biblical text and other text in \dev.

\section{Significance testing}
\label{sec:appendix-significance}

For LMs (\dev, and \test), contrasts were compared using mixed effects models implemented in R using the library \texttt{lme4}. Because the sentences translated are identical in all settings, we treat sentence ID as a random effect (repeated measures) and the addition of the lexicon and syntax augmentations as independent fixed effects, predicting the quality metric (BERT-Score F1). Single term deletions with likelihood ratio tests demonstrate that both lexicon and syntax augmentations improve translation quality as assessed by the metric for $p<0.0001$.

\section{Implementation and resources}

We ran our experiments with open-weighted models on Nvidia H100 GPUs. We maintain explicit requirements for reproducibility, and used a fixed random seed of $42$ in all our local runs for replicability.

With fixed random seed, we only need single run each for our model, setting, split that we report. Especially since there is no random effect commonly found in other training or augmentation. This was tested by running some of the experiments multiple times and checked for consistency.

\cref{tab:model-cards} lists the specific model cards for each of the models used.

\paragraph{Hyperparameters}

We use max tokens of $128$ for inference with both local and API models and use a fixed random seed. We used greedy decoding to avoid variance due to temperature. Data results and code are all documented in the repository for replicability and reproducibility.

Gridsearch parameters of the components are documented and available in paper (\cref{sec:appendix-lexgridsearch,sec:appendix-depgridsearch}) and available in the repository.

\draftversion{\section{Related Work}

Our work lies at the intersection of increasing consideration LLMs with instruction based MT~\cite{wmt25-instructions} both for evaluation and use, and increasing exploration of low-resource MT despite the many challenges~\cite{haddow-etal-2022-survey, court-elsner-2024-shortcomings}, specifically with additional language information. Here we discuss related works (many are also  background, see \cref{sec:background}) and contextualize out contributions.

Using additional resources (well) to improve MT is neither new not novel, however the way to do so can be specific to method and language pair. This can also be due to varying resource availability and accessibility for the languages of the world, espcially for low-resource languages~\cite{haddow-etal-2022-survey}. This has also been true for recent uses of LLMs for MT. 

Using grammar book to improve MT has been explored even in the recent years \cite{tanzer-2024-mtob, hus-anastasopoulos-2024-back}. Recently, \citet{pei-etal-2025-understanding} applied LLMs to MT in the low-resource language of Manchu~(ISO 639-3: \texttt{mnc})\footnote{\url{https://glottolog.org/resource/languoid/id/manc1252}}. They used morphological analyzer, dictionary information, and grammar book excerpts as augmentation, specifically looking to optimize the specific grammar book to use. Our work is the first to specifically and (also) successfully use explicit syntactic analysis to improve low-resource MT. We additional explore multiple operationalizations for adding this information. Hence contribution of work is three fold, first we provide novel approach of using syntactic analysis to improve in-context translation, second we explore multiple operationalization, and finally we do so for Coptic and provide both improved (ostraca) and fresh translation results for this low-resource language. 
}
\section{AI Tool Use}We used  \href{https://github.com/features/copilot}{GitHub Copilot} and \href{https://gemini.google.com/app}{Gemini} when developing the code for this paper.

\end{document}